\documentclass[letterpaper]{article} %
\usepackage{aaai2026}  %
\usepackage{times}  %
\usepackage{helvet}  %
\usepackage{courier}  %
\usepackage[hyphens]{url}  %
\usepackage{graphicx} %
\usepackage{natbib}  %
\usepackage{caption} %
\usepackage[utf8]{inputenc} %
\usepackage[T1]{fontenc}    %
\usepackage{url}            %
\usepackage{booktabs}       %
\usepackage{amsfonts}       %
\usepackage{nicefrac}       %
\usepackage{microtype}      %
\usepackage{xcolor}         %

\usepackage{mathtools}
\usepackage[frozencache]{minted}
\usepackage{algorithm}
\usepackage{algpseudocode}
\usepackage{float}

\usepackage{tikz}
\usetikzlibrary{arrows.meta,positioning,calc,fit,shapes,decorations.pathreplacing,shadows.blur}

\usepackage{siunitx}      %
\usepackage{multirow}     %
\usepackage{rotating}     %
\usepackage{graphicx}     %

\usepackage{chngcntr}

\nocopyright

\def\myTitle{Symmetry-Aware Transformer Training for Automated Planning} %
\title{\myTitle}

\author{%
  Markus Fritzsche\equalcontrib, Elliot Gestrin\equalcontrib, Jendrik Seipp \\ %
}

\affiliations{
    Linköping University, Sweden\\
    firstname.lastname@liu.se \\
  }

\definecolor{tblue}{RGB}{0,119,187}
\definecolor{tcyan}{RGB}{51,187,238}
\definecolor{tteal}{RGB}{0,153,136}
\definecolor{torange}{RGB}{238,119,51}
\definecolor{tred}{RGB}{204,51,17}
\definecolor{tmagenta}{RGB}{238,51,119}
\definecolor{tgray}{RGB}{187,187,187}

\definecolor{darkgreen}{rgb}{0.0, 0.2, 0.13}

\newcommand{\notes}[1]{}

\renewcommand{\cite}[1]{\citep[][]{#1}}
\newcommand{\egcite}[1]{\citep[e.g.,][]{#1}}
\newcommand{\inlinecite}[1]{\citet{#1}}

\newcommand{\define}[1]{#1}
\newcommand{\mypar}[1]{\smallskip\noindent\textbf{#1.}}

\newcommand{\token}[1]{\textit{#1}}
\newcommand{\pddl}[1]{\textit{#1}}

\newcommand{\pre}{\ensuremath{\textit{pre}}}

\newcommand{\add}{\ensuremath{\textit{add}}}
\newcommand{\del}{\ensuremath{\textit{del}}}

\newcommand{\states}{\ensuremath{\mathcal{S}}}
\newcommand{\predicates}{\ensuremath{\mathcal{P}}}

\newcommand{\objects}{\ensuremath{\mathcal{O}}}
\newcommand{\vocabulary}{\ensuremath{\mathcal{V}}}
\newcommand{\actionschemas}{\ensuremath{\mathcal{A}}}
\newcommand{\init}{\ensuremath{\mathcal{I}}}
\newcommand{\goal}{\ensuremath{\mathcal{G}}}

\DeclareMathOperator*{\argmax}{arg\,max}
\DeclareMathOperator*{\argmin}{arg\,min}

\newcommand{\tuple}[1]{\ensuremath{\langle #1 \rangle}}
\def\NA{\text{N/A}}

\usepackage{tabularx}

\newcommand{\predloss}{\ensuremath{L_\text{pred}}}
\newcommand{\attloss}{\ensuremath{L_\text{att}}}
\newcommand{\hidloss}{\ensuremath{L_\text{hid}}}

\begin{document}

\maketitle

\begin{abstract}
  While transformers excel in many settings, their application in the field of automated planning is limited.
  Prior work like PlanGPT, a state-of-the-art decoder-only transformer, struggles with extrapolation from easy to hard planning problems.
  This in turn stems from problem symmetries: planning tasks can be represented with arbitrary variable names that carry no meaning beyond being identifiers.
  This causes a combinatorial explosion of equivalent representations that pure transformers cannot efficiently learn from.
  We propose a novel contrastive learning objective to make transformers symmetry-aware and thereby compensate for their lack of inductive bias.
  Combining this with architectural improvements, we show that transformers can be efficiently trained for either plan-generation or heuristic-prediction.
  Our results across multiple planning domains demonstrate that our symmetry-aware training effectively and efficiently addresses the limitations of PlanGPT.
\end{abstract}

\noindent
\textbf{Code} --- \url{https://github.com/mrlab-ai/symmetry-transformers} \\
\noindent
\textbf{Data} \hspace{-1.1pt} --- \url{https://doi.org/10.5281/zenodo.17591697} \\
\noindent
\textbf{This work was accepted at AAAI 2026.}

\section{Introduction}

Transformer architectures~\cite{vaswani-et-al-nips2017} have revolutionized various fields, from natural language processing \egcite{tunstall-et-al-2022} to computer vision \egcite{zhai-et-al-cvpr2022}.
While leveraging transformers for automated planning---complex reasoning tasks that require sequential decisions---holds significant promise, transformers specifically trained for this purpose remain rare.
This is surprising, given that solving automated planning problems is inherently a sequence generation task: finding a sequence of actions that leads from the initial state to a goal state.

We posit that a key challenge lies in the inherent ambiguity of planning problem descriptions, combined with a lack of inductive bias in pure transformer architectures.
While pre-trained large language models have been explored for solving planning problems \egcite{plansformer,zeroShotPlanners,silver2024generalized,kambhampatiPositionLLMsCant2024,huang2025planning}, models trained from scratch for generating plans from structured planning inputs are much less common.
A notable exception is \emph{PlanGPT}~\cite{rossetti-et-al-icaps2024}, a state-of-the-art GPT-2-based transformer~\cite{radford-et-al-misc2019} trained to predict plans from formal problem descriptions.

\begin{figure*}[t]
    \centering
    \input{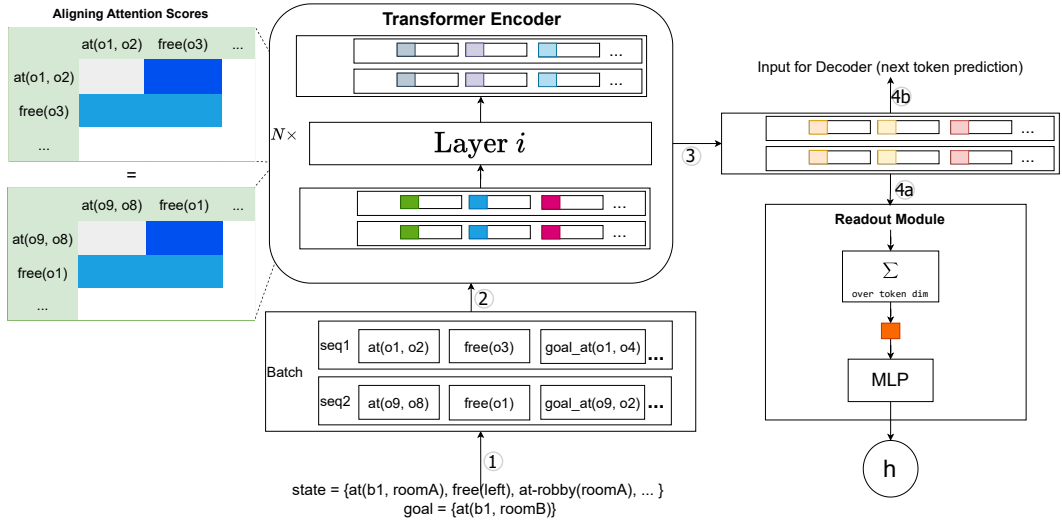}
    \caption{Illustration of our contrastive loss and the encoder part of our architecture.
        First, (1) we create two copies of the state and goal, each mapped into a token sequence, where each object is renamed to a random object token. While forwarding the sequences to the encoder (2), all attention scores and hidden states are collected for loss computations. The first objective is to align the attention scores for both samples in all attention-based modules (illustrated as equal attention scores). The second objective is to align fractions of the hidden states for all pairwise-corresponding tokens (illustrated by matching colors in the image). The output (3) can either be forwarded to the decoder 
        (4b, Figure~A.1 in the appendix) 
        for next-token prediction or to a readout module for heuristic computation (4a). When predicting heuristics, we sum the hidden state fractions used in the contrastive loss and pass them through an MLP to obtain the prediction.}

    \label{fig:dl_encoder}
\end{figure*}

However, PlanGPT faces several limitations, particularly concerning extrapolation to more challenging instances beyond its training set.
A primary reason for this is that the same planning problem can have a huge number of different input representations since 1) objects can be given arbitrary names, since they only serve as identifiers and carry no semantic meaning, and 2) the atoms of initial and goal states can be ordered arbitrarily.
Pure transformer architectures must implicitly learn to ignore the differences between such \emph{symmetric} planning problems, which is challenging and sample-inefficient.
Furthermore, relying on an explicitly learned positional encoding can hinder generalization, as the model may struggle with novel input sizes and unseen positions.%

We propose a symmetry-aware contrastive learning objective for transformers that explicitly accounts for the symmetries present in planning problems, guiding the model to learn representations that are equivariant to these symmetries (see Figure~\ref{fig:dl_encoder}).
While we focus on automated planning, this objective is applicable to any task using transformer tokens that primarily represent identifiers.
To address remaining symmetries in the input representation, we highlight the importance of architectural choices, especially the omission of explicit positional encodings.
Furthermore, we use a compositional tokenization scheme along with a transformer encoder, benefiting from its inherent permutation-equivariance.
We denote our architectures as \textbf{Sym}metry-Aware \textbf{T}ransformers (SymT).

We evaluate our approach in two different settings: \mbox{1) \emph{plan generation}} using an encoder-decoder ($\text{SymT}^{\text{ED}}$) and \mbox{2) \emph{heuristic prediction}} using an encoder-only transformer ($\text{SymT}^{\text{E}}$).
Our experiments show that our symmetry-aware contrastive learning objective and architectural innovations significantly improve planning performance compared to the PlanGPT baseline in three of the four evaluated domains, particularly in extrapolating to harder planning problems where our models find plans for many tasks that PlanGPT fails to solve.
These findings showcase the importance of explicitly addressing input symmetries to unlock the full potential of transformers for automated planning and general reasoning tasks that use variable names.

\section{Background}
\label{sec:background}

\mypar{Classical Planning}
We consider classical planning, where the world is fully-observable and actions are deterministic.
Formally, a (lifted STRIPS) planning problem~\cite{fikes-nilsson-aij1971} is a tuple $\tuple{\predicates, \objects, \actionschemas, \init, \goal}$. Here, $\predicates$ is a finite set of \define{predicates} that describe properties of the world (e.g., \pddl{at}, \pddl{in}, \pddl{connected}) and the arity of predicate $p \in \predicates$ is its number of arguments.
$\objects$ is a finite set of objects that exist in the world (e.g., \pddl{pkg1}, \pddl{truck3}, \pddl{loc2}).
Each object has a type (e.g., \pddl{Package}, \pddl{Truck}, \pddl{Location}).
Using predicates and objects, we form \define{(ground) atoms}, which are basic propositions about the world (e.g., \pddl{at(pkg1, loc2)}).
A \define{state} is a set of ground atoms that are true in a given situation and the initial state $\init$ describes which atoms are true initially.
We denote the set of all possible states by $\states$.
An atom can also contain \define{variables} (e.g., \pddl{at(?pkg, ?loc)}), which act as placeholders for objects.
Such \emph{lifted} atoms are used within action schemas $A \in \actionschemas$, which are templates for actions in the world.
Each action schema $A \in \actionschemas$ consists of a \define{precondition} $\pre(A)$, an \define{add list} $\add(A)$ and a \define{delete list} $\del(A)$.
The precondition $\pre(A)$ is a set of atoms that must be true in a state for the action to be applicable (e.g., \pddl{at(?pkg, ?loc)} and \pddl{at(?truck, ?loc)}).
The add list $\add(A)$ is a set of atoms that will be true after the action is applied (e.g., \pddl{in(?pkg, ?truck)}), and the delete list $\del(A)$ is a set of atoms that will no longer be true after the action is applied (e.g., \pddl{at(?pkg, ?loc)}).
We ground an action schema $A$ to obtain a ground action by replacing all variables in $\pre(A)$, $\add(A)$ and $\del(A)$ with objects from $\objects$, respecting the types of the variables and objects.
A ground action $a$ is \define{applicable} in state $s$ if $\pre(a) \subseteq s$, and applying it results in the \define{successor state} $s' = (s \setminus \del(a)) \cup \add(a)$.
The goal $\goal$ is the set of atoms that must be true simultaneously to solve the planning problem.
A sequence of ground actions is a \emph{plan} if it is applicable from $\init$ and reaches a goal state $s^*$ such that $\goal \subseteq s^*$.
Optimal planning is the task of finding a plan with minimum length.
Here, we focus on satisficing planning, where any plan is acceptable, but short plans are preferred.
Finally, a heuristic $h: \states \to {R}^+_0 \cup \infty$ is a function estimating the cost of reaching a goal state from a given state.

\mypar{Transformer Architectures for Sequence Generation}
Pure transformers~\cite{vaswani-et-al-nips2017} consist of an encoder and a decoder, each consisting of a stack of multiple layers.
The input to each layer is a sequence of token embeddings, and the output is a sequence of hidden state vectors.
Both types of layers consist of a multi-head attention (MHA) mechanism and a position-wise feed-forward network (MLP), with residual connections and optional layer normalization applied after each sub-layer.
Multi-head attention and self-attention (A) are defined as
\begin{align*}
    \alpha_{i}(Q, K) &= \text{softmax}\left(Q_{i} K_{i}^T / \sqrt{d_k}\right) \\
    \text{A}_{i}(Q, K, V) &= \alpha_{i}(Q, K) V_{i} \\
    \text{MHA}(Q, K, V) &= (\text{A}_{1}(Q, K, V) | ... | \text{A}_{L}(Q, K, V)) W_O,
\end{align*}
where $W_O \in \mathbb{R}^{d\times d}$ is a learned projection combining the $L$ attention heads, and $|$ denotes concatenation.
$Q, K, V \in \mathbb{R}^{N \times d}$ are the query, key, and value matrices, with $Q_i, K_i, V_i \in \mathbb{R}^{N \times d_k}$ the $i$-th head and $d_k = d / L$.

An encoder layer transforms the input $X^{(i)} \in \mathbb{R}^{N \times d}$ into the output $X^{(i+1)} \in \mathbb{R}^{N \times d}$ as follows:
\begin{align*}
    Q^{(i)} &= X^{(i)} W_Q^{(i)} ~~~ K^{(i)} = X^{(i)} W_K^{(i)} ~~~ V^{(i)} = X^{(i)} W_V^{(i)} \\
    X_{\text{att}}^{(i)} &= X^{(i)} + \text{MHA}^{(i)}\left(Q^{(i)}, K^{(i)}, V^{(i)}\right) \\
    X^{(i+1)} &= X_{\text{att}}^{(i)} + \text{MLP}^{(i)}\left(X_{\text{att}}^{(i)}\right)
\end{align*}
Here, $W_Q^{(i)}, W_K^{(i)}, W_V^{(i)} \in \mathbb{R}^{d \times d}$ are learned weight matrices specific for the $i$-th layer.
We omit layer normalization for simplicity.
Optionally, a bias term can be added to the Q, K, and V matrices, but we omit it here for simplicity.

We let $E$ denote the output of the last encoder layer and $Y \in \mathbb{R}^{M \times d}$ the decoder input sequence.
A decoder layer transforms the input $Y^{(i)}$ into the output $Y^{(i+1)}$ as follows:
\begin{align*}
    Q_Y^{(i)} &= Y^{(i)} W_{Q,1}^{(i)} ~~~ K_Y^{(i)} = Y^{(i)} W_{K, 1}^{(i)} ~~~ V_Y^{(i)} = Y^{(i)} W_{V, 1}^{(i)} \\
    Y_{\text{att}}^{(i)} &= Y^{(i)} + \text{MHA}^{(i)}\left(Q_Y^{(i)}, K_Y^{(i)}, V_Y^{(i)}\right) \\
    Q_{Y_{\text{att}}}^{(i)} &= Y_{\text{att}}^{(i)} W_{Q, 2}^{(i)} \quad K_E^{(i)} = E W_{K, 2}^{(i)} \quad V_E^{(i)} = E W_{V, 2}^{(i)} \\
    \tilde{Y}^{(i)} &= Y_{\text{att}}^{(i)} + \text{MHA}^{(i)}\left(Q_{Y_{\text{att}}}^{(i)}, K_E^{(i)}, V_E^{(i)}\right) \\
    Y^{(i+1)} &= \tilde{Y}^{(i)} + \text{MLP}^{(i)}\left(\tilde{Y}^{(i)}\right)
\end{align*}

The first MHA module in the decoder masks out future tokens using a triangular mask, ensuring that the model can only attend to previous tokens in the sequence.
A final MLP predicts the probability of the next token in the sequence, generating the output sequence one token at a time, referred to as \define{autoregressive generation}.
Typically, the decoder is trained using the cross-entropy loss between the predicted token probabilities and the true next token in the sequence.
Decoder-only variants, such as GPT-based models \egcite{radford-et-al-misc2018,rossetti-et-al-icaps2024}, focus solely on this sequential generation process and use only self-attention.

\mypar{Transformers for Automated Planning}
The primary target for Deep Learning in planning has been the prediction of heuristics, where the model is trained to predict the estimated cost of reaching the goal from a given state~\egcite{toyer-et-al-jair2020,stahlberg-et-al-icaps2022,chen-et-al-icaps2024}.
While transformers can be used for predicting heuristics, most work has focused on the sequence-to-sequence setting, where the model is trained to predict a plan (sequence of actions) given a planning problem, i.e., a sequence of state and goal atoms \cite{plansformer,rossetti-et-al-icaps2024}.
Unlike architectures with strong built-in assumptions about local connectivity such as graph neural networks (GNNs), transformers possess limited inductive bias regarding sequence structure, relying heavily on data to learn dependencies.
Handling sequence order explicitly requires positional information, commonly via positional encodings, as the attention mechanism itself is permutation-invariant.
This data-driven nature and reliance on learned positional signals pose a challenge, particularly when input sequences can exhibit symmetries or structural variations not explicitly encoded, as is the case in planning problems.

\mypar{PlanGPT}
\emph{PlanGPT}~\cite{rossetti-et-al-icaps2024} is a decoder-only transformer architecture, trained to predict a plan for a given planning problem.
The model processes the planning problem by segmenting it into an input token sequence.
The input begins with a \token{<start>} token, followed by tokens representing the initial state, where each atom is decomposed into predicate and argument tokens (e.g., atom \pddl{at(truck5, loc3)} becomes the token sequence \token{at}, \token{truck5}, \token{loc3}).
A \token{<goal>} token separates the state from the goal description, tokenized similarly.
The input sequence concludes with an \token{<action>} token, signaling the start of the plan output.
The output sequence comprises tokens representing the predicted actions, also segmented into action name and argument tokens, until the end-of-sequence token \token{<EOS>} is generated.
An example input-output sequence for a successfully-solved simplified Logistics tasks is: \textit{<start> at truck5 loc1 at pkg1 loc1 \dots <goal> at pkg1 loc2 <action> load truck5 pkg1 loc1 move truck5 loc1 loc2 unload truck5 pkg1 loc2 <EOS>}.
To handle symmetries induced by arbitrary object names, PlanGPT randomizes object names based on their types from a fixed vocabulary, e.g., \token{truck5} becomes \token{truck2}.
Training is supervised using a standard next-token prediction objective, namely cross-entropy loss.
The authors use a training dataset that contains multiple (suboptimal) plans for each planning problem, allowing the model to learn from diverse solutions.

\section{Limitations of PlanGPT}

PlanGPT is trained solely to minimize the next-token prediction loss.
While this objective can in principle capture symmetries, the combinatorial explosion of input representations leads to poor sample efficiency in practice.

\mypar{Limitation 1: Object Assignment Equivariance}
Object names in planning are arbitrary and not intended to carry additional semantics.
However, PlanGPT uses a fixed vocabulary of object names that are randomly assigned to objects in the planning problem.
Assume a single object type with $|\objects|$ objects in the training instance and $|\vocabulary|$ object names in vocabulary $\vocabulary$ with $|\vocabulary| \geq |\objects|$.
Then there are $\frac{|\vocabulary|!}{(|\vocabulary| - |\objects|)!}$ different assignments of objects to vocabulary names.
For example, a task with $|\objects| = 8$ objects and $|\vocabulary| = 8$ names yields $\frac{8!}{(8 - 8)!} = 40320$ different assignments, all describing the same planning task.
The model must learn to generalize across all assignments to predict plans equivariant to these name assignments, which is challenging without additional objectives.

\mypar{Limitation 2: Leaking Information in Object Names}
To mitigate Limitation~1, PlanGPT does not randomize object names of all types.
In the Visitall domain, for instance, objects of type \pddl{Location} encode grid coordinates, e.g., \pddl{loc-x1-y2} corresponds to coordinate $\tuple{1, 2}$.
Instead of randomizing these names, PlanGPT keeps the human-intended coordinates, which the model can memorize, preventing generalization to locations not seen during training.
We demonstrate this in Appendix~B1.

\mypar{Limitation 3: Atom Order Invariance}
The order of atoms in the initial and goal states of a planning problem is arbitrary.
Assume there are $|\init|$ atoms in the state and $|\goal|$ atoms in the goal description.
Then there are $|\init|!$ different atom orders for the same state and $|\goal|!$ different orders for the goal,
yielding $|\init|! \cdot |\goal|!$ equivalent representations for each object assignment from Limitation~1.
PlanGPT is a decoder-only model that uses learned positional encodings to capture token order in the input sequence.
To make atom order consistent across training instances, PlanGPT pre-sorts the atoms in the state and goal descriptions before passing them to the model.
Nevertheless, the model may overfit to particular atoms being overrepresented in specific positions, which can hinder generalization to larger instances where atoms appear in different positions.

\mypar{Limitation 4: Learned Positional Encodings}
In contrast to state and goal atoms, the order of actions in a plan matters.
Since transformer layers are permutation-equivariant functions,\footnote{If we reorder the input sequence of tokens according to some permutation, the output sequence will be the original output sequence reordered by the same permutation.} PlanGPT uses learned absolute positional encodings~\cite{radford-et-al-misc2019} to capture the order of input tokens.
This is problematic for state and goal tokens due to Limitation~3.
Moreover, for any token type, learned positional encodings imply that embeddings for unseen positions are not learned, preventing generalization to harder planning problems with more objects, which typically require longer plans.

\section{Symmetry-Aware Training}
\label{sec:aware-training}
This section introduces our contrastive training objective and architecture choices for overcoming the limitations identified for PlanGPT.
An overview of our objective and encoder-only architecture for heuristics is shown in Figure~\ref{fig:dl_encoder},
while our encoder-decoder architecture for plan generation is shown in Figure~A.1 (Appendix~A).

\mypar{Addressing Atom Order Symmetries with Encoders \mbox{(Limitation 3)}} %
To directly address atom order symmetries, we use an encoder-only architecture for heuristic prediction and an encoder-decoder architecture for plan generation.
The encoder processes the state and goal atoms, while the decoder predicts the action sequence forming the plan.
By omitting positional encodings in the encoder, its processing is inherently permutation-equivariant with respect to the order of input atom tokens.
This makes its output representation independent of their arbitrary order, yielding a $|\init|! \cdot |\goal|!$ reduction of the input space.

PlanGPT tokenizes each atom into several tokens (predicate and arguments), each mapped to an individual learnable embedding and requiring an ordering.
As this is incompatible with an encoder without positional encodings, we instead encode each atom as a single embedding.
Let $E[\text{token}]$ denote the learnable embedding vector for a token.
For an atom $p(o_1, \dots, o_n)$ grounded from predicate $p$ and objects $o_1, \dots, o_n$, we concatenate $E[p]$, $E[o_1], \dots, E[o_n]$, and $m-n$ padding token embeddings to obtain a vector of fixed length $|E[\text{token}]| \cdot (1 + m)$, where $m$ is the maximum predicate arity in the domain.
We then pass this vector through a single linear layer with parameters $\mathbf{W}$ and $\mathbf{b}$ to produce the atom embedding $T_{p(o_1, \dots, o_n)}$:
\begin{align*}
    &T_{p(o_1, \dots, o_n)} = \mathbf{W} (E[p] | E[o_1] | \dots | E[o_n] | E[\text{pad}] | \dots) + \mathbf{b}
\end{align*}
This linear transformation maps the concatenated embedding to the dimensionality of the atom embeddings used by the encoder.\footnote{Directly encoding atoms with unique tokens, i.e., $E[p(o_1, \dots, \allowbreak o_n)]$, would induce a massive vocabulary that scales poorly in the number of supported objects, making it likely that some atoms are rarely seen during training and causing poor generalization.}
We further avoid splitting the input into state and goal subsequences by introducing new goal-predicates for every predicate.
For example, the goal atom \pddl{at(truck3, loc2)} becomes \pddl{goal\_at(truck3, loc2)}.
By converting all atoms into these atom-level embeddings and operating without positional encodings, the encoder produces a representation that is guaranteed to be permutation-equivariant in the order of input atom tokens, addressing Limitation~3.

\mypar{Addressing Positional Encodings in the Decoder \mbox{(Limitation 4)}} %
Predicting heuristic values only requires an order-invariant state representation and is therefore suitable for an encoder-only architecture without positional encodings.
In contrast, the order of actions in a plan is essential and must be preserved.
As noted in Limitation~4, learned positional encodings can struggle to generalize to unseen positions and thus limit extrapolation.
To mitigate this issue in the decoder, we omit positional encodings entirely.
This technique, \emph{NoPE}, has shown successful length generalization in various sequence-to-sequence tasks, including reasoning problems~\cite{kazemnejad-et-al-neurips2023}.

\mypar{Contrastive Training for Object Name Equivariance (Limitations 1 and 2)}
To obtain robust object name assignment equivariance \mbox{(Limitation 1)}, architectural changes alone are insufficient.
We therefore introduce a contrastive training objective that encourages the model to learn representations and processing patterns equivariant to arbitrary object name assignments.

The core idea is to train on pairs of symmetric planning problems.
For a given problem, we generate two encoder input sequences, $X$ and $X'$, that share the same underlying state and goal structure but differ only in object names.
We similarly generate two corresponding decoder output sequences, $Y$ and $Y'$, which are the plans predicted for $X$ and $X'$, respectively.
Since $X$ and $X'$ represent the same planning problem, the model should process them in a way that reflects this equivalence.
We propose two renaming modes:
\textit{Rename-One}, where object names in $X$ are fixed across training instances and object names in $X'$ are randomized, and \textit{Rename-Both}, where object names in both $X$ and $X'$ are randomized.
We then rename the objects in $Y$ and $Y'$ accordingly.
Preliminary experiments showed that Rename-One works better for predicting heuristics, whereas Rename-Both is better for plan generation.
We encourage equivalent behavior across both sequences using two complementary contrastive losses.

The first, the attention loss $\attloss$, is motivated by the intuition that if the model encodes the same algorithm regardless of names, the attention scores between corresponding parts of the input must match across layers and heads (Figure~\ref{fig:dl_encoder}).
We define $\attloss$ as:
\begin{align*}
    L_{\text{att}} &= \frac{1}{B} \sum_{\alpha \in \textit{Att}} \sum_{i = 1}^{\text{\#rows}(\alpha)} \sum_{j = 1}^{\text{\#cols}(\alpha)} (\alpha_{i, j} - \alpha'_{i, j})^2
\end{align*}
Here, $B$ denotes the batch size and $\textit{Att}$ the set of all attention modules across layers and heads.
$\alpha$ and $\alpha'$ are the attention scores produced by the transformer for $(X, Y)$ and $(X', Y')$, respectively.

The second component, the hidden state loss $\hidloss$, encourages similarity directly in the learned token representations.
It is based on the idea that the hidden state vector for a token should encode problem-structure features that are independent of the specific object names.
We compute $\hidloss$ as:
\begin{align*}
    \hidloss &= \frac{1}{B} \sum_{H \in \{X, Y\}} \sum_{l=1}^{\text{\#layers}(H)}  ( H^{(l)}_{[:d_k]} - H'^{(l)}_{[:d_k]})^2
\end{align*}
Here, $H^{(l)}$ denotes the hidden states after the $l$:th encoder layer when $H = X$ and the $l$:th decoder layer when $H = Y$, and $[{:}d_k]$ denotes a projection selecting the first $d_k$ dimensions of these hidden states.

We combine these with the standard prediction loss:
    $L = w_1 \cdot \predloss + w_2 \cdot \attloss + w_3 \cdot \hidloss$.
Here, $\predloss$ is the next-token prediction loss (cross-entropy for plan generation, MSE for heuristic prediction), and $w_1$, $w_2$, and $w_3$ are hyperparameters controlling the importance of each component.
We set all $w_i$ to 1 in our experiments.
Minimizing this objective incentivizes the model to predict correct actions or heuristic values while learning attention patterns and hidden state features that are equivariant to object name variations.
For predicting heuristics with our encoder-only architecture, we feed $\sum_N X^{(N)}_{[:d_k]}$ instead of $\sum_N X^{(N)}$ to the readout MLP to obtain the final heuristic prediction (Figure~\ref{fig:dl_encoder}).
We additionally address Limitation~2 by always fully randomizing object names.

\mypar{Summary of Our Architectures}
We introduce two symmetry-aware transformers: $\text{SymT}^{\text{E}}$, an encoder-only architecture for heuristic prediction, and $\text{SymT}^{\text{ED}}$, an encoder-decoder architecture for plan generation.
Inspired by the GNNs of \inlinecite{stahlberg-et-al-icaps2022}, both share weights across layers to reduce the number of parameters, see Table~\ref{tab:parameters}.
This is strictly necessary; otherwise $L_{\text{hid}}$ would collapse to a trivial solution for slices of hidden states in non-final layers.
Both models use the contrastive loss described above to ignore object names.
Thus, our heuristic model is a permutation-equivariant, shared-weight encoder, and our plan generation model extends it with a shared-weight decoder using NoPE positional encodings for length generalization.

\begin{table}[t]
    \centering
    \begin{tabular}{lccc}
        & PlanGPT & $\text{SymT}^{\text{E}}$ & $\text{SymT}^{\text{ED}}$ \\
        \midrule
        \# Parameters & 117M & 7M & 16M
    \end{tabular}
    \caption{Parameter counts for PlanGPT and our architectures.}
    \label{tab:parameters}
\end{table}

\section{Plan Generation}
\label{sec:decoding}

There are many ways to use transformer models to generate plans, differing in how much they rely on the model versus search techniques such as backtracking or cycle detection.
To assess the performance of the model rather than the search strategy, we consider three minimally supportive plan generation strategies.

\mypar{Greedy Plan Generation}
At each step, we autoregressively select the token with the highest predicted probability and stop when an end-of-sequence token is generated.

\mypar{Applicability-Filtered Plan Generation}
This strategy again autoregressively selects the most probable token, but first masks the probabilities so that only tokens leading to an applicable action are considered.
For example, if the last two tokens generated are \token{move} and \token{roomA}, then the model can only generate a token corresponding to a room reachable from \token{roomA} (e.g., \token{roomB}).
Plan generation stops when the simulated state satisfies the goal.
This strategy dominates the greedy plan generation strategy.

\mypar{Regrounding Applicability-Filtered Plan Generation}
This strategy filters applicable actions as above.
After each action is generated, it is applied to update the state, and the model's input sequence is reset to this new state.
Essentially, we alternate between predicting (partial) plans of length one and applying them.
Again, we stop when the found state satisfies the goal.
This strategy does not dominate either of the two above.

\mypar{Greedy Heuristic Guidance}
For the heuristic-prediction model, we generate plans by computing the heuristic value of each successor state and greedily selecting the action that leads to the state with the lowest heuristic value.
This process is repeated until reaching a goal state. %

For each strategy, we also stop plan generation early if the model generates more than 500 tokens.
Pseudo-code for all strategies is in Appendix~D.

\section{Experiments, Results and Limitations}
\label{sec:experiments}

We now evaluate our symmetry-aware training, focusing on training efficiency and extrapolation capabilities.

  \setlength{\tabcolsep}{2pt}
  \begin{table*}[tb]
    \renewcommand{\arraystretch}{1.1}
    \centering
    \resizebox{\textwidth}{!}{
      \newcolumntype{Y}{>{\centering\arraybackslash}X}
 \begin{tabularx}{\textwidth}{clYYYYYYY}
 & & \multicolumn{3}{c}{PlanGPT - Decoder (baseline)} & \multicolumn{1}{c}{$\text{SymT}^\text{E}$ (ours)} & \multicolumn{3}{c}{$\text{SymT}^\text{ED}$ (ours)} \\
 \cmidrule(lr){3-5} \cmidrule(lr){6-6} \cmidrule(lr){7-9}
 & & greedy & applicable & regrounding & greedy & greedy & applicable & regrounding \\
 \midrule
 \multirow{3}{*}{\rotatebox[origin=c]{90}{Blocks}} & validation & ~\,.00$\pm$.00 & ~\,.00$\pm$.00 & ~\,.00$\pm$.00 & \textbf{1.00$\pm$.00} & \textbf{1.00$\pm$.00} & \textbf{1.00$\pm$.00} & ~\,.00$\pm$.00 \\
 & interpolation & ~\,.56$\pm$.16 & ~\,.56$\pm$.16 & ~\,.00$\pm$.00 & \textbf{1.00$\pm$.00} & \textbf{1.00$\pm$.00} & \textbf{1.00$\pm$.00} & \textbf{1.00$\pm$.00} \\
 & extrapolation & ~\,.00$\pm$.00 & ~\,.00$\pm$.00 & ~\,.00$\pm$.00 & ~\,.05$\pm$.07 & ~\,.07$\pm$.02 & \textbf{~\,.13$\pm$.05} & ~\,.00$\pm$.00 \\
 \midrule
 \multirow{3}{*}{\rotatebox[origin=c]{90}{Gripper}} & validation & ~\,.00$\pm$.00 & ~\,.00$\pm$.00 & ~\,.00$\pm$.00 & \textbf{1.00$\pm$.00} & ~\,.17$\pm$.24 & \textbf{1.00$\pm$.00} & \textbf{1.00$\pm$.00} \\
 & interpolation & ~\,.00$\pm$.00 & ~\,.44$\pm$.16 & ~\,.00$\pm$.00 & ~\,.89$\pm$.16 & ~\,.67$\pm$.00 & \textbf{1.00$\pm$.00} & \textbf{1.00$\pm$.00} \\
 & extrapolation & ~\,.00$\pm$.00 & ~\,.00$\pm$.00 & ~\,.00$\pm$.00 & ~\,.02$\pm$.03 & ~\,.00$\pm$.00 & ~\,.15$\pm$.06 & \textbf{~\,.79$\pm$.16} \\
 \midrule
 \multirow{3}{*}{\rotatebox[origin=c]{90}{Visitall}} & validation & ~\,.00$\pm$.00 & ~\,.14$\pm$.12 & ~\,.00$\pm$.00 & \textbf{1.00$\pm$.00} & ~\,.33$\pm$.09 & ~\,.93$\pm$.04 & ~\,.99$\pm$.02 \\
 & interpolation & ~\,.05$\pm$.04 & ~\,.67$\pm$.18 & ~\,.41$\pm$.22 & \textbf{1.00$\pm$.00} & ~\,.87$\pm$.01 & ~\,.99$\pm$.01 & \textbf{1.00$\pm$.00} \\
 & extrapolation & ~\,.00$\pm$.00 & ~\,.02$\pm$.02 & ~\,.00$\pm$.00 & ~\,.42$\pm$.11 & ~\,.00$\pm$.00 & ~\,.15$\pm$.05 & \textbf{~\,.64$\pm$.12} \\
 \midrule
 \multirow{3}{*}{\rotatebox[origin=c]{90}{Logistics}} & validation & ~\,.00$\pm$.00 & \textbf{~\,.08$\pm$.12} & ~\,.00$\pm$.00 & ~\,.00$\pm$.00 & ~\,.00$\pm$.00 & ~\,.00$\pm$.00 & ~\,.00$\pm$.00 \\
 & interpolation & ~\,.07$\pm$.05 & \textbf{~\,.44$\pm$.09} & ~\,.19$\pm$.14 & ~\,.11$\pm$.00 & ~\,.22$\pm$.31 & ~\,.26$\pm$.29 & ~\,.22$\pm$.31 \\
 & extrapolation & \textbf{~\,.00$\pm$.00} & \textbf{~\,.00$\pm$.00} & \textbf{~\,.00$\pm$.00} & \textbf{~\,.00$\pm$.00} & \textbf{~\,.00$\pm$.00} & \textbf{~\,.00$\pm$.00} & \textbf{~\,.00$\pm$.00} \\
 \bottomrule
 \end{tabularx}
    }
    \caption{Normalized coverage scores $ [(\mu \pm \sigma)] $ for all architectures when decoding with the various generation strategies. A coverage score of 1.00 means that the configuration solves all tasks. The best results are highlighted in bold.
}
    \label{tab:coverage_split_base_woCL}
  \end{table*}

\mypar{Experimental Setup}
Since the benchmarks used by PlanGPT are unsuitable for our purposes, due to significant training/test set overlap (see Appendix~B2) and a lack of dedicated extrapolation data, we generated our own datasets
for four widely used planning domains with characteristics relevant to PlanGPT's limitations: \textit{Blocksworld}, \textit{Gripper}, \textit{Visitall} and \textit{Logistics}.
In Blocksworld, a set of blocks in multiple towers must be stacked in a specific order into a single tower.
In Gripper, a robot with two grippers must move a set of balls from one room to another.
In Visitall, an agent must visit all cells in a grid.
In Logistics, a set of trucks and airplanes must transport packages from their initial locations to their goal locations.
Appendix~C details the domains, their PDDL representations, and the size and distribution of problems for each domain.

We train models for each combination of domain and architecture on an NVIDIA A100 GPU with 12 hours of training per model; Appendix~E gives further details.
We then evaluate them using all plan generation strategies from Section~\ref{sec:decoding} on three sets of problems: validation (slightly larger than training and used for checkpoint selection), interpolation (comparable but distinct sizes), and extrapolation (larger than validation).
For robustness, we use three random seeds per configuration.
For the PlanGPT baseline, we avoid leaking object names by always randomizing all object names.
We additionally perform ablation studies to analyze the impact of our contrastive loss (adding it to PlanGPT and removing it from our models) and to examine the effect of the number of objects on the contrastive loss; see Appendix~E2 and E3, respectively.
Performance is primarily evaluated based on \emph{coverage}, i.e., the percentage of problems for which a valid plan is generated, and \emph{plan quality}, i.e., the length of the generated plans compared to best known plan lengths.

\mypar{Results -- Coverage}
Coverage is shown in Table~\ref{tab:coverage_split_base_woCL}.
PlanGPT can solve instances similar in size to the training problems (interpolation), but fails entirely on larger instances regardless of decoding strategy.
In contrast, both our architectures solve larger instances to varying degrees.
If regrounding extrapolates well for a domain, applicability-filtered plan generation tends to perform worse, and vice versa, with regrounding generally preferable.
$\text{SymT}^{\text{ED}}$ also consistently outperforms $\text{SymT}^{\text{E}}$ in coverage.
No approach extrapolates in Logistics, which aligns with known GNN results~\cite{stahlberg-et-al-icaps2022} based on expressiveness limitations.
Logistics is also the only domain where PlanGPT outperforms our models, likely due to its larger number of parameters allowing better fit to the training data.
Even there, however, our $\text{SymT}^{\text{ED}}$ model outperforms PlanGPT in both greedy and regrounding settings.

\mypar{Results -- Plan Quality}
Appendix~E2 reports plan quality tables, where we see that the generated plans are often optimal or close to optimal.
This likely stems from training exclusively on optimal plans, encouraging the models to reproduce such policies.
Although not our main focus, we also observe that the $\text{SymT}^{\text{ED}}$ finds shorter plans in seconds than the classical planner LAMA~\cite{richter-westphal-jair2010} does in two hours for several problems (we use LAMA to obtain reference plan lengths), indicating that the learned policies are practically useful.

\mypar{Results -- Training Stability}
Training with the contrastive objective yields lower validation losses, but the coverage gains are most pronounced for the heuristic prediction model.
Following \inlinecite{abbe-et-al-neurips2024}, we hypothesize that this is because predicting goal distances requires considering all input tokens, whereas autoregressive next-action prediction does not.
For example, in Gripper predicting the next action only requires considering a few balls, while the goal distance depends on all balls.

Across many domains, losses plateau for $\text{SymT}^{\text{E}}$ models without the contrastive term, while models with it continue to improve.
We also observed frequent training instabilities without the contrastive loss for $\text{SymT}^{\text{E}}$ and $\text{SymT}^{\text{ED}}$, leading to sudden loss divergence; standard stabilization techniques such as gradient clipping or reduced learning rates did not help.
Across all trained models (including ablations and hyperparameter tuning), only a single model diverged with the contrastive loss, compared to 19 without it.
We discuss the link between the contrastive objective and training stability in Appendix~E.\\[-1em]

\mypar{Limitations}
While we outperform PlanGPT on extrapolation, our models still have clear limitations.
They extrapolate to some extent under greedy plan generation, but the applicability-filtered and regrounding strategies perform better, indicating that the learned algorithm is not perfect.
In the hardest domain, Logistics, we cannot extrapolate to larger instances at all.
Extrapolation success likely depends on several factors, such as the choice of positional encodings, and further work is needed to understand their impact on length generalization.
Another limitation is that all transformer models, including PlanGPT, rely on a fixed-size vocabulary, so problems with more objects than vocabulary entries cannot be solved.
As we show in Appendix~E3, however, our contrastive loss allows stable training with much larger vocabularies, making it viable to choose a large vocabulary ahead of training as we do in our experiments.

\section{Related Work}

Learning-based approaches for automated planning increasingly complement traditional symbolic methods, aiming to improve performance, handle more complex domains, and enable faster planning.

\mypar{Learning for Planning with Graph Neural Networks}
GNNs~\cite{scarselli-et-al-ieeenn2009} are the predominant deep learning architecture for learning directly from structured planning problem representations~\cite{stahlberg-et-al-icaps2022,stahlberg-et-al-kr2022,chen-thiebaux-neurips2024,stahlberg-et-al-aaai2025,horcik-et-al-aaai2025}.
Their inductive bias towards processing graph-structured data, where nodes and edges can represent atoms and their relationships, aligns well with the relational nature of planning problems.
However, standard GNNs are theoretically bounded by their expressive power, aligning with the 1-dimensional Weisfeiler--Lehman (WL) test~\cite{xu-et-al-iclr2019}, and may struggle to distinguish WL-indistinguishable graph structures.
Approaches like higher-order GNNs~\cite{stahlberg-et-al-aaai2025} have been explored to lift this expressiveness barrier for planning tasks.
Despite their success, GNNs have not been exploited for generating plans in the context of next-token generation.

\mypar{Transformers for Planning and Structured Data}
Applying transformer architectures directly to structured planning problem inputs is less common than using GNNs.
Notable examples include \emph{PlanGPT}~\cite{rossetti-et-al-icaps2024}, which treats planning as a sequence generation task, and \emph{Dualformer}~\cite{su-et-al-iclr2025}, which trains transformers on search traces for navigation and puzzle tasks.
A key challenge for these sequence-based models is extrapolation to larger instances.
Transformers, particularly those relying on learned absolute positional encodings, are often limited in their ability to generalize to sequence lengths significantly different from those seen during training~\cite{su-et-al-iclr2025,rossetti-et-al-icaps2024}.
Methods like iterative self-training~\cite{lee-et-al-icml2025} show that transformers can be trained to extrapolate on some tasks, including graph-encoded mazes, but typically require complex multi-round training strategies.
To the best of our knowledge, training transformers for out-of-distribution extrapolation on STRIPS planning problems \emph{without} such multi-stage fine-tuning remains an open challenge.

\mypar{Large Language Models for Planning}
Recent research increasingly leverages pre-trained large language models (LLMs) for solving planning problems.
Some work fine-tunes LLMs specifically for planning tasks~\cite{plansformer} or uses them to generate programs that solve planning problems~\cite{silver2024generalized,katz-et-al-neurips2024}.
Other strategies employ LLMs to generate plans directly from natural language problem descriptions~\cite{zeroShotPlanners}, with increasingly intricate prompting strategies~\cite{treeOfThoughts,sel-et-al-icml2024,gestrin-et-al-icaps2024wshaxp,sel-et-al-iclr2025} and sometimes assisted by external tools~\cite{kambhampatiPositionLLMsCant2024}, or to translate natural language into PDDL tasks that can be solved with off-the-shelf planners~\cite{llmPlusP,oswald2024large,huang2025planning}.
These approaches capitalize on the extensive knowledge embedded within LLMs and their inherent reasoning capabilities.
In contrast, our work focuses on training models directly on structured planning data (like PDDL tokens) to obtain robust and generalizable plan generation models.

\mypar{Equivariant and Invariant Learning}
Training models to be equivariant or invariant to specific input transformations is a fundamental theme in deep learning, notably in computer vision (e.g., CNNs for translation invariance) and GNNs (permutation-equivariance).
For planning problems, symmetries such as the arbitrary assignment of object names introduce complex permutations that ideal learned models should handle.
Some work explores architectural modifications to achieve equivariance for structured data, e.g., Graphormer~\cite{ying-et-al-neurips2021}.
Relevant to object name symmetries, \emph{Renamer}~\cite{ankner-et-al-neurips2023wsmlsys} proposes transformer architectural changes for semantics-preserving variable renaming.
Our work complements such architectural efforts by exploring training objectives.

\mypar{Objectives on Intermediate Representations}
Standard training often focuses on the final output loss (e.g., prediction error), but some methods introduce auxiliary losses on intermediate layer outputs, such as hidden states or attention weights, to encourage desired properties in the learned representations.
We found no prior work that specifically introduces a contrastive learning objective applied directly to attention scores or targeted hidden state projections \emph{between symmetrically equivalent inputs} with the explicit goal of inducing object name assignment equivariance in a transformer model.
However, there is work that uses losses on attention for other purposes~\cite{patro-et-al-wacv2021}.

\section{Conclusions and Future Work}

By introducing a novel contrastive training objective designed to encourage object name equivariance in the learned representations, combined with architectural choices to address planning-related symmetries, we showed improved inter- and extrapolation performance compared to the state-of-the-art PlanGPT baseline across multiple planning domains.
However, despite significant gains over the baseline, our model is still unable to consistently generate valid plans for problems \emph{significantly} larger than those encountered during training.
This outcome reinforces that achieving robust, out-of-distribution extrapolation on complex symbolic reasoning tasks remains a substantial challenge for existing transformer architectures and training paradigms.

Future work should identify the architectural or representational bottlenecks that limit generalization.
Also, addressing the practical limitation of a fixed vocabulary size by exploring representation learning techniques independent of token vocabularies is crucial for tackling very large tasks.

\section*{Acknowledgments}

This work was partially supported by the Wallenberg AI, Autonomous Systems and Software Program (WASP) funded by the Knut and Alice Wallenberg Foundation.
The computations were enabled by resources provided by the National Academic Infrastructure for Supercomputing in Sweden (NAISS), partially funded by the Swedish Research Council through grant agreement no. 2022-06725.

\bibliography{abbrv-short,literatur,nl2plan,crossref-short}

\appendix
\onecolumn
\setcounter{secnumdepth}{2}
\renewcommand\thefigure{\thesection.\arabic{figure}}
\counterwithin{figure}{section}
\renewcommand\thetable{\thesection.\arabic{table}}
\counterwithin{table}{section}
\renewcommand\thesubsection{\thesection \arabic{subsection}}

\begin{center}
    {\huge  \textbf{\myTitle \\[0.5em] Appendix \\[1em]}}
\end{center}

\section{Decoder Architecture}
\label{app:decoder}
\begin{figure*}[h!]
    \centering
    \input{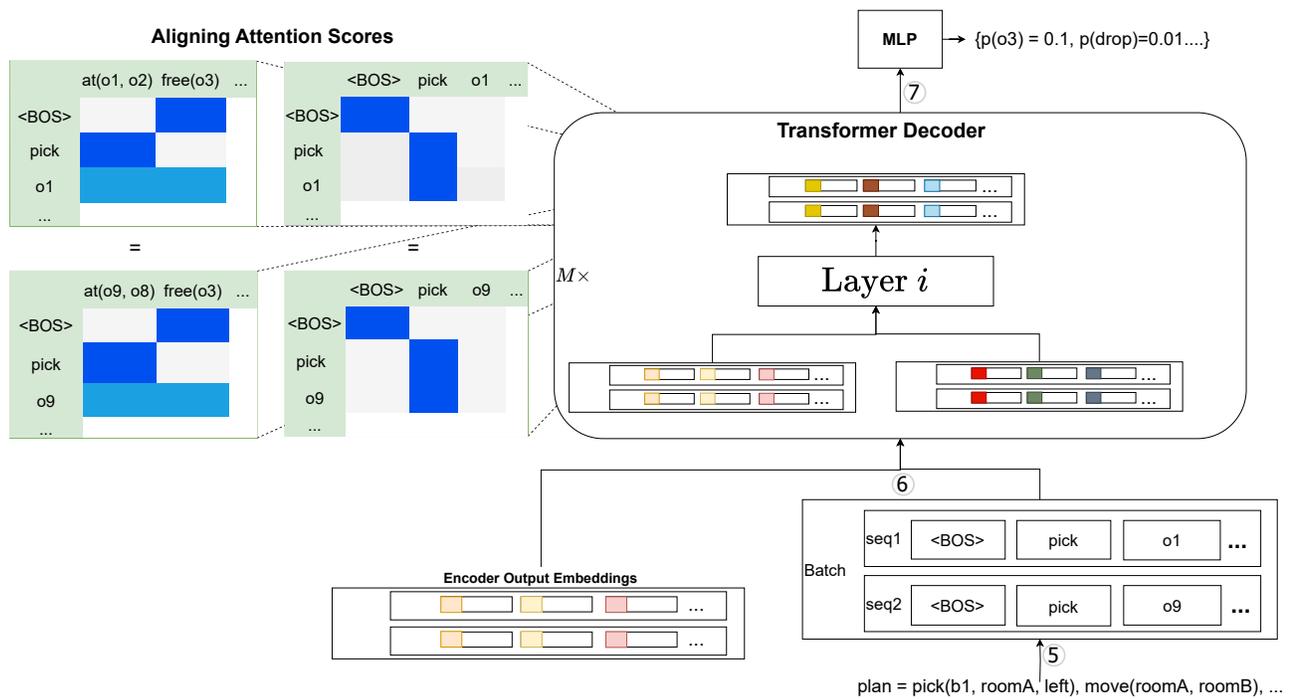}
    \caption{Illustration of the decoder part of our architecture. First, we copy the input plan, and rename all objects using the same renaming scheme as in the encoder and map both copies into sequences. While forwarding the sequences and encoder output to the decoder (6), all attention scores and hidden states are collected and aligned equally as explained in Figure~\ref{fig:dl_encoder}. The decoder output is then used to predict the next token in the plan (7).}
    \label{fig:dl_decoder}
\end{figure*}

Figure \ref{fig:dl_decoder} illustrates the decoder part of our architecture and the associated contrastive loss.

\vfill
\vfill

\clearpage
\section{PlanGPT Limitations}
\label{app:plangpt-remarks}

\subsection{Data Leakage through Object Names}
\label{app:plangpt-remarks:renaming}

To illustrate the impact of object renaming on PlanGPT, we evaluated their published models on 3 domains (Blocksworld, Visitall and Floortile) with and without object renaming. Note that this also used their original training and test data. The results are shown in Table~\ref{tab:plangpt-renaming}. We note that all information remains in the problem and that classical planning approaches would be unaffected by this renaming, and that our models would be far more resilient.

Blocksworld is a domain in which the object names have no semantic meaning. As such, PlanGPT is unaffected by the renaming.

Visitall is a domain in which the object names are semantically meaningful. The object names are based on the coordinates of the locations, e.g., \pddl{loc-x1-y2} stands for the coordinate $\tuple{1, 2}$ in the grid. As such, renaming the locations to random names leads to a complete failure of the model, showing that it acts primarily on the inferred meaning of the object names rather than the supplied predicates.

Floortile is a domain which has two types, robots and tiles. We also note that the coverage we found without renaming is significantly lower than the reported value. We are unaware of the reason for this discrepancy, but hypothesize that the authors accidentally published the wrong model checkpoints. Both the robots and tiles are semantically meaningful, as the robots' names correlate to the color they carry and the tiles again are directly correlated to their coordinates. While it manages to solve $.05\%$ of the problems when renaming only the robots, this is still a massive drop in performance.

We only evaluated this for these three domains and only for a single random renaming each.

These results highlight how PlanGPT often utilizes leaked information in the object names to solve problems. This is a significant limitation, as it means that the model is unable to generalize to unseen objects or locations, which is a common requirement in planning problems. Our architectures on the other hand, have to infer this information from the available predicates since all the objects are anonymized. This makes them act in the way classical planners do and the way a user would expect a planning model to behave.

\begin{table}[h]
    \centering
    \caption{PlanGPT's performance on the original training data with and without object renaming. Reported values are those reported by the authors in \inlinecite{rossetti-et-al-icaps2024}. The other columns show the performance of the models when rerunning their evaluation without object renaming (None) and with object renaming of the stated type(s). The coverage values are deterministic for a fixed renaming. We only evaluated a single random renaming for each category.}
    \small
    \begin{tabular}{@{}l ccc ccc ccccc@{}}
        & \multicolumn{3}{c}{Blocksworld} & \multicolumn{3}{c}{Visitall} & \multicolumn{5}{c}{Floortile} \\
        \cmidrule(lr){2-4} \cmidrule(lr){5-7} \cmidrule(lr){8-12}
        & Reported & None & Blocks & Reported & None & Locations & Reported & None & Robots & Tiles & Both \\
        \midrule
        \multirow{1}{*}{Coverage [\%]} & 99.5 & 99.55 & 99.36 & 94 & 93.80 & 0 & 94.4 & 44.73 & 0.05 & 0 & 0 \\
    \end{tabular}
    \label{tab:plangpt-renaming}
\end{table}

\subsection{Training and Test Set Overlap}
\label{app:plangpt-remarks:train-test-overlap}
We discovered that PlanGPT's training and test sets overlap in the Visitall domain to a certain degree.
Of 6231 test problem files, 1931 identical problems were found in the training set as well.
Furthermore, of the 2419 unique initial states in the test set, 2418 were also found in the training set.
For that reason and the fact that no extrapolation data is included, we opted to train on our own datasets.
We did not evaluate this for any domain other than Visitall.

\newpage
\section{PDDL Domain Descriptions}
\label{sec:appendix:domains}

In this section, we provide high-level descriptions of the planning domains used in our experiments and their PDDL representations. In addition, we provide the sizes of the training, validation, interpolation and extrapolation sets for each domain in Table~\ref{tab:dataset_sizes}.
During training, we fully expand each training instance and sample from the complete state space. As such, our approach mitigates the risk of only seeing certain parts of the state space while also heavily reducing the number of problems that need to be used for training. The larger number of training problems in Visitall than other domains is due to the fact that the number of unique states in any given Visitall problem is noticeably smaller than in Blocksworld and Logistics. Consequently, the total number of training states thereof lies between Blocksworld and Logistics. On the contrary, in Gripper, all problems with the same number of balls are equivalent under the expansion and as such we only included one per training size. In comparison, PlanGPT training used 63 000 problems per domain for training \cite{rossetti-et-al-icaps2024}. See Appendix~\ref{app:experiments} for further details on the training procedure and sampling. To generate the instances, we either used commonly available domain generation scripts or wrote our own. These are available in our repository.

\begin{table}[h]
    \setlength{\tabcolsep}{4.4pt}
    \small
    \centering
    \caption{Problem sizes of the planning problems from the four benchmark subsets. The problems in the training set are chosen to allow exhaustive exploration of their state spaces. The problems in the extrapolation set are chosen to be large enough to allow for meaningful evaluation of the models' extrapolation capabilities.}
    \begin{tabular}{@{}lcccccc@{}}
        Domain & Parameter & \# Training Instances & Training Sizes & Validation Sizes & Interpolation Sizes & Extrapolation Sizes \\
        \midrule
        Blocksworld & \#blocks & 9 & 4, 6, 7 & 8 & 5 & 9, 10, \dots, 17 \\
        Gripper & \#balls & 4 & 2, 4, 6, 8 & 9, 10 & 3, 5, 7 & 12, 14, \dots, 42 \\
        Logistics & \#goals & 12 & 1, 3, 5 & 6 & 2, 4 & 7, 8, \dots, 15 \\
        Visitall & \#cells & 207 & 1, 3, 4, 6, 10, 11, 12, 14, 16 & 18, 20 & 2, 5, 8, 9, 15 & 24, 25, \dots, 121  \\
    \end{tabular}
    \label{tab:dataset_sizes}
\end{table}

\subsection{Blocksworld}
\label{sec:appendix:blocks}

A Blocksworld problem consists of a set of blocks that can be picked up and stacked on top of each other or put on the table.
The goal is to stack all blocks in a single stack in a given order.
Blocksworld problems are NP-hard to solve optimally~\cite{gupta-nau-aij1992}, but can be solved in at most twice the minimum number of steps using a simple polynomial policy that puts all blocks on the table and then stacks them in the goal order~\cite{selman-kr1994}.

\begin{minted}{text}
(define (domain blocks)
    (:requirements :strips)
    (:predicates (on ?x ?y)
                 (ontable ?x)
                 (clear ?x)
                 (handempty)
                 (holding ?x))

    (:action pick-up
        :parameters (?x)
        :precondition (and (clear ?x)
                           (ontable ?x)
                           (handempty))
        :effect (and (not (ontable ?x))
                     (not (clear ?x))
                     (not (handempty))
                     (holding ?x)))

    (:action put-down
        :parameters (?x)
        :precondition (holding ?x)
        :effect (and (not (holding ?x))
                     (clear ?x)
                     (handempty)
                     (ontable ?x)))

    (:action stack
        :parameters (?x ?y)
        :precondition (and (holding ?x)
                           (clear ?y))
        :effect (and (not (holding ?x))
                     (not (clear ?y))
                     (clear ?x)
                     (handempty)
                     (on ?x ?y)))

    (:action unstack
        :parameters (?x ?y)
        :precondition (and (on ?x ?y)
                           (clear ?x)
                           (handempty))
        :effect (and (holding ?x)
                     (clear ?y)
                     (not (clear ?x))
                     (not (handempty))
                     (not (on ?x ?y)))))
\end{minted}

\subsection{Gripper}
\label{sec:appendix:gripper}

A Gripper problem consists of a robot with two gripper arms that can  pick up and drop balls and move between two rooms, $A$ and $B$.
The goal is to move all $n$ balls from room $A$ to $B$.
The family of Gripper problems can be solved optimally in polynomial time~\cite{helmert-aij2003}.
One optimal policy is to:
\begin{itemize}
    \item Always move to room $A$ when both grippers are empty.
    \item Once in room $A$, pick up as many balls as possible.
    \item If in $A$ and unable to pick more balls, move to room $B$.
    \item In room $B$, drop all balls.
    \item Repeat until all balls are in room $B$.
\end{itemize}
While this domain is simple, extrapolation to larger instances is not trivial for learning algorithms.
This is particularly true for transformers as the model must learn to keep track of the ball positions while dealing with plan token positions.

\begin{minted}{text}
(define (domain gripper)
  (:requirements :strips :negative-preconditions)
  (:predicates
    (room ?r)
    (ball ?b)
    (gripper ?g)
    (at-robby ?r)
    (at ?b ?r)
    (free ?g)
    (carry ?o ?g)
  )

  (:action move
    :parameters (?from ?to)
    :precondition (and
      (room ?from)
      (room ?to)
      (at-robby ?from)
    )
    :effect (and
      (at-robby ?to)
      (not (at-robby ?from))
    )
  )

  (:action pick
    :parameters (?obj ?room ?gripper)
    :precondition (and
      (ball ?obj)
      (room ?room)
      (gripper ?gripper)
      (at ?obj ?room)
      (at-robby ?room)
      (free ?gripper)
    )
    :effect (and
      (carry ?obj ?gripper)
      (not (at ?obj ?room))
      (not (free ?gripper))
    )
  )

  (:action drop
    :parameters (?obj ?room ?gripper)
    :precondition (and
      (ball ?obj)
      (room ?room)
      (gripper ?gripper)
      (carry ?obj ?gripper)
      (at-robby ?room)
    )
    :effect (and
      (at ?obj ?room)
      (free ?gripper)
      (not (carry ?obj ?gripper))
    )
  )
)
\end{minted}

\subsection{Logistics}
\label{sec:appendix:logistics}

A Logistics problem consists of a set of trucks, locations, airports, cities, airplanes and packages.
The goal is to move all packages from their initial locations to their goal locations.
Cities have locations, some of which are airports.
Trucks can move packages between locations within a city, while airplanes can move packages between airports.
The Logistics domain is more complex than Gripper, and even state-of-the-art deep learning-based planners struggle to extrapolate to larger instances due to their expressive limitations~\cite{stahlberg-et-al-kr2023}.
Finding optimal plans for Logistics is NP-complete~\cite{helmert-aij2003}, but suboptimal plans can be generated in polynomial time using a greedy algorithm that delivers one package at a time~\cite{helmert-et-al-ecai2006}.
While we do not expect our model to extrapolate to larger instances, we can still compare the interpolation performance to instances of similar size as the training instances.
Furthermore, compared to Gripper, the Logistics domain has many different object types and PlanGPT assigns object names based on the object type.
We use Logistics problems to compare the impact of this object assignment method compared to a fully-random assignment of object names.

\begin{minted}{text}
(define (domain logistics)
  (:requirements :strips)
  (:predicates
    (package ?obj)
    (truck ?truck)
    (airplane ?airplane)
    (airport ?airport)
    (location ?loc)
    (in-city ?obj ?city)
    (city ?city)
    (at ?obj ?loc)
    (in ?obj1 ?obj2))

(:action load-truck
  :parameters (?obj ?truck ?loc)
  :precondition
   (and (package ?obj) (truck ?truck) (location ?loc)
        (at ?truck ?loc) (at ?obj ?loc))
  :effect
   (and (not (at ?obj ?loc)) (in ?obj ?truck)))

(:action load-airplane
  :parameters (?obj ?airplane ?loc)
  :precondition
   (and (package ?obj) (airplane ?airplane) (location ?loc)
        (at ?obj ?loc) (at ?airplane ?loc))
  :effect
   (and (not (at ?obj ?loc)) (in ?obj ?airplane)))

(:action unload-truck
  :parameters (?obj ?truck ?loc)
  :precondition
   (and (package ?obj) (truck ?truck) (location ?loc)
        (at ?truck ?loc) (in ?obj ?truck))
  :effect
   (and (not (in ?obj ?truck)) (at ?obj ?loc)))

(:action unload-airplane
  :parameters (?obj ?airplane ?loc)
  :precondition
   (and (package ?obj) (airplane ?airplane) (location ?loc)
        (in ?obj ?airplane) (at ?airplane ?loc))
  :effect
   (and (not (in ?obj ?airplane)) (at ?obj ?loc)))

(:action drive-truck
  :parameters (?truck ?loc-from ?loc-to ?city)
  :precondition
   (and (truck ?truck) (location ?loc-from) (location ?loc-to) (city ?city)
        (at ?truck ?loc-from) (in-city ?loc-from ?city) (in-city ?loc-to ?city))
  :effect
   (and (not (at ?truck ?loc-from)) (at ?truck ?loc-to)))

(:action fly-airplane
  :parameters (?airplane ?loc-from ?loc-to)
  :precondition
   (and (airplane ?airplane) (airport ?loc-from) (airport ?loc-to)
        (at ?airplane ?loc-from))
  :effect
   (and (not (at ?airplane ?loc-from)) (at ?airplane ?loc-to)))
)
\end{minted}

\subsection{Visitall}
\label{sec:appendix:visitall}

A problem in Visitall consists of a set of locations in a four-connected rectangular grid and an agent that has to visit all cells.
Finding shortest paths that visit all cells in arbitrary rectangular grids is NP-complete~\cite{itai-et-al-sicomp1982}.
However, there is a simple suboptimal policy that incurs at most twice the number of steps of an optimal plan: walk to a corner, then move in snake-like motion through all rows of the grid.
We include the Visitall domain because even slight increases in the number of objects, i.e., the locations, make problems much harder to solve.
Furthermore, PlanGPT assigns location names not arbitrarily as in the other domains, but based on the coordinates of the locations, which adds additional semantic meaning to the object names [personal communication with the authors of PlanGPT].
This implies that the model does not need to learn the meaning of the \textit{adjacent} predicate, which is used to describe the adjacency of locations.
Additionally, the model is fundamentally limited to the locations in the training set, as it cannot generalize to unseen locations.

\begin{minted}{text}
(define (domain grid-visit-all)
  (:requirements :strips :negative-preconditions)
  (:predicates
    (place ?x)
    (connected ?x ?y)
    (at-robot ?x)
    (visited ?x)
  )

  (:action move
    :parameters (?curpos ?nextpos)
    :precondition (and
      (place ?curpos)
      (place ?nextpos)
      (at-robot ?curpos)
      (connected ?curpos ?nextpos)
    )
    :effect (and
      (at-robot ?nextpos)
      (not (at-robot ?curpos))
      (visited ?nextpos)
    )
  )
)
\end{minted}

\clearpage
\section{Decoding Strategies}
\label{app:plan-algorithms}
\subsection{Greedy Plan Generation}

\begin{algorithm}[H]
\caption{Greedy Plan Generation. Starting from the encoding of the planning problem, autoregressively generate a plan by appending the token with the highest predicted probability at each step.}
\label{alg:greedy_prediction_limited}
\begin{algorithmic}[1]
\Require Planning problem $P = \langle \predicates, \objects, \mathcal{A}, \init, \goal \rangle$
\Require Encoder function $\text{Encode}(\objects, \init, \goal)$
\Require Transformer model $M$
\Require Beginning-of-sequence token $\texttt{<BOS>}$
\Require End-of-sequence token $\texttt{<EOS>}$
\Require Maximum plan length $N$
\Ensure Predicted action sequence $\text{Plan}$

\State $E \leftarrow \text{Encode}(\predicates, \objects, \init, \goal)$
\State $\text{Plan} \leftarrow \tuple{\texttt{<BOS>}}$

\While{$\text{Plan}$ does not end with $\texttt{<EOS>}$ and length($\text{Plan}$) $< N$}
    \State $\text{InputSequence} \leftarrow E \oplus \text{Plan}$
    \State $\text{Predictions} \leftarrow M(\text{InputSequence})$
    \State $\text{NextToken} \leftarrow \argmax_{t} \text{Predictions}$
    \State append $\text{NextToken}$ to $\text{Plan}$
\EndWhile

\State \Return $\text{Plan}$
\end{algorithmic}
\end{algorithm}
\clearpage

\subsection{Applicability-Filtered Plan Generation}
\label{app:applicability-filtered-plan-generation}

\begin{algorithm}[H]
    \caption{Applicability-Filtered Plan Generation. At each step, select the highest probability action token. Before appending it to the plan, simulate its effect in the planning environment. If the action is invalid from the current state, it is masked, and the model's probabilities for other actions are re-evaluated. Plan generation stops when the simulated state reaches the goal.}
    \label{alg:applicability-filtered-plan-generation}
    \begin{algorithmic}[1]
    \Require Planning problem $P = \langle \predicates, \objects, \mathcal{A}, \init, \goal \rangle$
    \Require Encoder function $\text{Encode}(\predicates, \objects, \init, \goal)$
    \Require Transformer model $M$
    \Require Beginning-of-sequence token $\texttt{<BOS>}$
    \Require End-of-sequence token $\texttt{<EOS>}$
    \Require Maximum plan length $N$
    \Require Function $\text{ApplyPlan}(\text{InitialState}, \text{Plan})$ returns the final state after executing the plan
    \Require Function $\text{IsGoalState}(\text{State}, \text{Goal})$ returns True iff State satisfies Goal
    \Require Simulator function $\text{MaskInvalidTokens}(\text{State}, \text{ActionSeq}, \text{Predictions})$ sets probabilities of tokens corresponding to inapplicable actions in State to zero
    \Ensure Predicted action sequence $\text{Plan}$

    \State $E \leftarrow \text{Encode}(\predicates, \objects, \init, \goal)$
    \State $\text{Plan} \leftarrow \tuple{\texttt{<BOS>}}$

    \While{not $\text{IsGoalState}(\text{CurrentState}, \goal)$ and length($\text{Plan}$) $< N$}
        \State $\text{CurrentState} \leftarrow \text{ApplyPlan}(\init, \text{Plan})$
        \State $\text{InputSequence} \leftarrow E \oplus \text{Plan}$
        \State $\text{Predictions} \leftarrow M(\text{InputSequence})$
        \State $\text{MaskedPredictions} \leftarrow \text{MaskInvalidTokens}(\text{CurrentState}, \text{Plan}, \text{Predictions})$
        \State $\text{NextToken} \leftarrow \argmax_{t} \text{MaskedPredictions}[-1, t]$
        \State append $\text{NextToken}$ to $\text{Plan}$
    \EndWhile

    \State \Return $\text{Plan}$
    \end{algorithmic}
    \end{algorithm}
\clearpage

\subsection{Regrounding Applicability-Filtered Plan Generation}
\label{app:pure-simulation-plan-generation}

\begin{algorithm}[H]
\caption{Regrounding Applicability-Filtered Plan Generation. At each step, the model's output probabilities are filtered to only include valid actions applicable in the current simulated state. Among these valid actions, the model selects the one with the highest predicted probability autoregressively. The provided model state is then updated by simulating the selected action, and the process continues until a goal state is reached.}
\label{alg:greedy_single_action_token_limit}
\begin{algorithmic}[1]
\Require Planning problem $P = \langle \predicates, \objects, \mathcal{A}, \init, \goal \rangle$
\Require Encoder function $\text{Encode}(\predicates, \objects, \init, \goal)$
\Require Transformer model $M$
\Require Beginning-of-sequence token $\texttt{<BOS>}$
\Require End-of-sequence token $\texttt{<EOS>}$
\Require Maximum number of tokens $N$
\Require Function $\text{IsAction}(\text{TokenSeq})$ tests if the token sequence is a complete action
\Require Function $\text{ApplyAction}(\text{State}, \text{Action})$ returns the state reached by executing Action in State
\Require Function $\text{IsGoalState}(\text{State}, \text{Goal})$ returns True iff State satisfies Goal
\Require Simulator function $\text{MaskInvalidTokens}(\text{State}, \text{TokenSeq}, \text{Predictions})$ sets probabilities of tokens corresponding to inapplicable actions in State to zero
\Ensure Executed action sequence Plan

\State $E \leftarrow \text{Encode}( \predicates, \objects, \init, \goal)$
\State $\text{CurrentState} \leftarrow \init$
\State $\text{Plan} \leftarrow \tuple{}$
\State $\text{CurrentTokenSeq}$ $\leftarrow \tuple{\texttt{<BOS>}}$

\While{not $\text{IsGoalState}(\text{CurrentState}, \goal)$ and length($\text{Plan}$) $< N$}

    \State $\text{InputSequence} \leftarrow E \oplus \text{CurrentTokenSeq}$
    \State $\text{Predictions} \leftarrow M(\text{InputSequence})$

    \State $\text{MaskedPredictions} \leftarrow \text{MaskInvalidTokens}(\text{CurrentState}, \text{CurrentTokenSeq}, \text{Predictions})$

    \State $\text{NextToken} \leftarrow \argmax_{t} \text{MaskedPredictions}$

    \State append NextToken to CurrentTokenSeq
    \If{$\text{IsAction}(\text{CurrentTokenSeq})$}
        \State $\text{CurrentState} \leftarrow \text{ApplyAction}(\text{CurrentState}, \text{CurrentTokenSeq})$
        \State append CurrentTokenSeq to Plan
        \State $E \leftarrow \text{Encode}( \predicates, \objects, \text{CurrentState}, \goal)$
        \State $\text{CurrentTokenSeq} \leftarrow \tuple{\texttt{<BOS>}}$

    \EndIf

\EndWhile

\State \Return Plan
\end{algorithmic}
\end{algorithm}
\clearpage

\subsection{Greedy Heuristic Guidance}
\label{app:greedy-heuristic-plan-generation}

\begin{algorithm}[H]
    \caption{Greedy Heuristic Guidance. Starting from the encoding of the planning problem, autoregressively generate a plan by appending the applicable action causing the lowest estimated number of steps to solve the planning problem.}
    \label{alg:greedy_heuristic_limited}
    \begin{algorithmic}[1]
    \Require Planning problem $P = \langle \predicates, \objects, \mathcal{A}, \init, \goal \rangle$
    \Require Encoder function $\text{Encode}(\objects, \init, \goal)$
    \Require Transformer model $M$ estimates the number of steps to solve a planning problem
    \Require Maximum number of actions $N$
    \Require Function $\text{IsGoalState}(\text{State}, \text{Goal})$ returns True iff State satisfies Goal
    \Require Function $\text{GetApplicableActions}(\text{State})$ returns the actions applicable in the given state
    \Ensure Executed action sequence Plan

    \State $\text{CurrentState} \leftarrow \init$
    \State $\text{Plan} \leftarrow \tuple{}$
    \While{not $\text{IsGoalState}(\text{CurrentState}, \goal)$ and length($\text{Plan}$) $< N$}
        \State $E \leftarrow \text{Encode}(\predicates, \objects, \text{CurrentState}, \goal)$
        \State $\text{ApplicableActions} \leftarrow \text{GetApplicableActions}(\text{CurrentState})$
        \State $\text{Action} \leftarrow \argmin_{a \in \text{ApplicableActions}} M(\text{Encode}(\predicates, \objects, \text{ApplyAction}(\text{CurrentState}, a), \goal))$
        \State $\text{CurrentState} \leftarrow \text{ApplyAction}(\text{CurrentState}, \text{Action})$
        \State Append Action to Plan
    \EndWhile
    \State \Return Plan
    \end{algorithmic}
\end{algorithm}

\clearpage
\section{Experiments}
\label{app:experiments}
For PlanGPT as a baseline, we use the hyperparameters and architecture as described in the original paper \cite{rossetti-et-al-icaps2024}, though sometimes augmented with our contrastive loss.

Our encoder and encoder-decoder models consist of 12 layers (in each, encoder and decoder), with a hidden dimension of 768, and 12 attention heads.
These are the equivalent as those the PlanGPT variant.
However, we implement shared weights between all encoder and decoder layers and we do not use any layer normalization. For positional encoding, we use NoPE~\cite{kazemnejad-et-al-neurips2023}, which is a learned positional encoding.

We train these models using the AdamW optimizer~\cite{loshchilov-hutter-iclr2019} together with a linear warmup schedule and subsequent cosine decay, with:
\begin{itemize}
    \item Weight decay: $0.1$
    \item $\beta_1$: $0.9$
    \item $\beta_2$: $0.999$
    \item Maximum iterations for schedules: $500000$
\end{itemize}
To decide on the remainder of the optimizer hyperparameters, we performed a hyperparameter search, as described in Section~\ref{app:encoder-decoder-hyperparams}.
Parameters for the contrastive losses, such as weights $w_1$, $w_2$ and $w_3$ are set to $1.0$, and the projection size $k$, is fixed at $32$ across all domains.
This value for $k$ was chosen since GNNs have been shown to only require $32$-dimensional representations~\egcite{stahlberg-et-al-icaps2022,stahlberg-et-al-kr2022, stahlberg-et-al-kr2023}.

The heuristic-prediction model (encoder) uses mean-squared error (MSE) loss while the plan-generation models (decoder and encoder-decoder) use cross-entropy loss.
For the contrastive loss, we use MSE across all architectures.

\begin{figure}[h]
    \centering
    \includegraphics[width=0.8\textwidth]{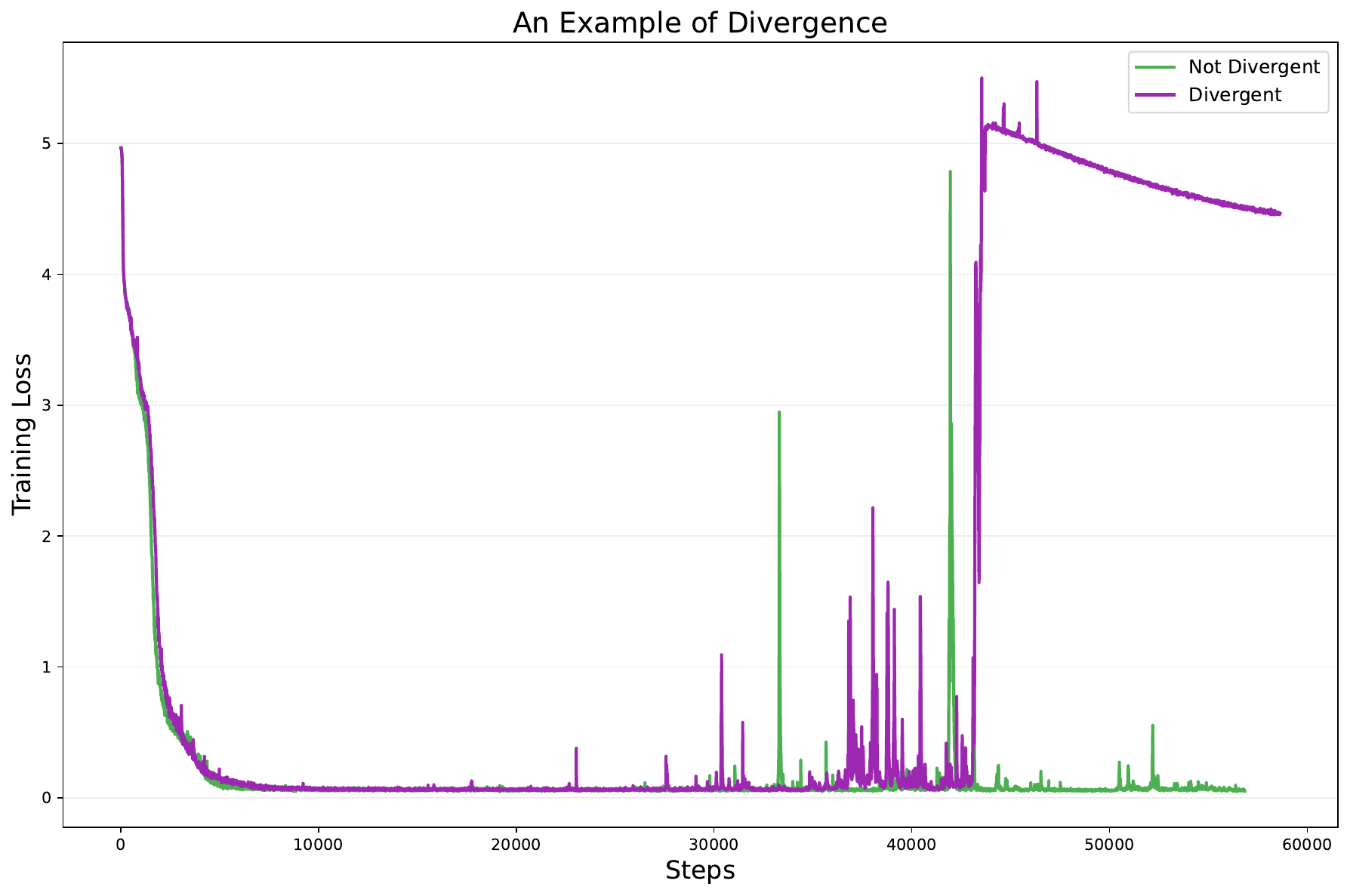}
    \caption{An example of model divergence. One line (purple) shows a model that diverged, suddenly spiking and stabilizing at a high plateau. The other (green) shows a model that did not diverge, while the loss spiked it also quickly returned to low values. The models shown are from the main experiment, trained on the Blocksworld domain without contrastive loss. When applying our contrastive loss during the main experiments, we did not observe neither any divergences nor any spikes. }
    \label{fig:divergence-example}
\end{figure}

A model under training was considered to diverge if its training loss suddenly spiked and stabilized at a high plateau, see Figure~\ref{fig:divergence-example} for an example, or became NaN.

For training, the PDDL instances are such that the state space can be exhaustively explored allowing for sampling optimal plans, a common practice in classical planning data generation~\cite{stahlberg-et-al-kr2022}.
Whenever we require a state-plan pair, we start by sampling a random problem from the training set.
Following this, we sample a random goal distance which occurs in this problem.
Then, we sample a state at this goal distance from which we lastly sample an optimal plan.
All samples are done uniformly at random.
For learning heuristics, we sample in the same way, but compute the optimal plan length $h^*$ instead of an optimal plan.
This approach helps guarantee diverse training data and avoids bias towards specific types of optimal plans.
Renaming is then applied to these samples.
Note that when batch sizes are reported below, they refer to the original number of samples before any renaming is applied. As such, the effective batch size is twice the reported batch size, as each sample is duplicated and renamed once.

Possibly, use of suboptimal or close-to-optimal plans could lead to better performance, as the model would have increased exposure to recoveries from mistakes -- though, likely at the cost of plan quality. However, this would require further ablation studies (e.g. degree of suboptimality). As such, we focus on optimal plans in this work.

Each model was trained on a single NVIDIA A100 GPU with 40GB of memory, 32GB of RAM and an AMD Epyc 9534 CPU on an internal cluster.

The number of test instances for interpolation and extrapolation is shown in Table~\ref{tab:testset_sizes}.
\begin{table}[h]
    \setlength{\tabcolsep}{4.4pt}
    \centering
    \caption{Number of instances in the interpolation and extrapolation test sets.}
    \begin{tabular}{@{}lccccc@{}}
        Domain & Interpolation & Extrapolation \\
        \midrule
        Blocksworld & 3 & 20 \\
        Gripper & 3 & 16 \\
        Logistics & 9 & 18 \\
        Visitall & 37 & 219  \\
    \end{tabular}
    \label{tab:testset_sizes}
\end{table}

\subsection{Hyperparameter Search}
\label{app:encoder-decoder-hyperparams}
We performed a hyperparameter search for the remaining optimization hyperparameters for the encoder-decoder architecture.
We selected the following hyperparameters for the search:
\begin{itemize}
    \item Warmup Steps: Number of warmup steps for the learning rate scheduler.
    \item Learning Rate: The initial learning rate for the optimizer.
    \item Minimum LR: The minimum learning rate for the optimizer.
    \item Dropout: The dropout rate used in the model.
\end{itemize}
For each hyperparameter, we chose a base value. We then fixed all but one hyperparameter to these base values and varied the remaining one. The base values were as follows:
\begin{itemize}
    \item Warmup Steps: 2000
    \item Learning Rate: 1.5e-4
    \item Minimum LR: 1e-7
    \item Dropout: 0
\end{itemize}
The considered values for each hyperparameter were:
\begin{itemize}
    \item Warmup Steps: 1000, 2000, 5000
    \item Learning Rate: 6e-4, 1.5e-4, 1e-4, 5e-5
    \item Minimum LR: 1e-6, 1e-7, 1e-8
    \item Dropout: 0, 0.1
\end{itemize}
For each combination of hyperparameters, we train one model with our loss and one without it.
Each model was trained with a timeout of 10 hours and up to a maximum of $40$ epochs, each consisting of $20000$ sampled states and plans, and a batch size of 32.
The same seed was used across all runs.
The maximum number of objects was set to $30$.

The results are reported in Table~\ref{tab:raw_hyperparam_encoder_decoder} and summarized in Table~\ref{tab:summarized_hyperparam_encoder_decoder}.
Note that all except one model reached the maximum number of epochs within the time limit, the exception reached epoch $37$.
The hyperparameters chosen are shown in Table~\ref{tab:summarized_hyperparam_encoder_decoder}.
Additionally, we report the number of runs and divergences for each choice of hyperparameters and note that the use of our contrastive loss led to a far lower divergence rate, showing that it increases the stability of the training process.

In total, we trained 24 models using 191.5 hours of GPU time for this experiment.

\newcommand{\tVal}{Yes}
\newcommand{\fVal}{No}

\begin{table}[htbp]
    \centering
    \caption{Results from hyperparameter search for the encoder-decoder model. These are summarized in Table~\ref{tab:summarized_hyperparam_encoder_decoder}.}
    \footnotesize
    \renewcommand{\arraystretch}{1.3}
    \begin{tabular}{@{} c
                    S[table-format=4.0, round-precision=0]
                    S[table-format=1.1e-1, round-precision=1] @{\hspace{10pt}}
                    S[table-format=1.0e-1, round-precision=0]
                    S[table-format=1.1]
                    c
                    S[table-format=1.4, round-precision=4]
                    S[table-format=1.4, round-precision=4]
                    @{}}
    \textbf{Contrastive Loss} & \textbf{Warmup} & \textbf{LR} & \textbf{Min LR} & \textbf{Dropout} & \textbf{Diverged} & \textbf{Train Loss} & \textbf{Validation Loss} \\
    \midrule
        \fVal & 1000 & 1.5e-4 & 1e-7 & 0 & \fVal & 0.05545411 & 0.13280071 \\
        \fVal & 2000 & 5e-5 & 1e-7 & 0 & \fVal & 0.06059282 & 0.11628938 \\
        \fVal & 2000 & 1e-4 & 1e-7 & 0 & \fVal & 0.04576498 & 0.10872728 \\
        \fVal & 2000 & 1.5e-4 & 1e-8 & 0 & \tVal & 3.00802207 & 3.01802325 \\
        \fVal & 2000 & 1.5e-4 & 1e-7 & 0 & \tVal & NaN & NaN \\
        \fVal & 2000 & 1.5e-4 & 1e-7 & 0 & \tVal & NaN & NaN \\
        \fVal & 2000 & 1.5e-4 & 1e-7 & 0 & \tVal & NaN & NaN \\
        \fVal & 2000 & 1.5e-4 & 1e-7 & 0 & \tVal & NaN & NaN \\
        \fVal & 2000 & 1.5e-4 & 1e-7 & 0.1 & \fVal & 0.05322025 & 0.11205934 \\
        \fVal & 2000 & 1.5e-4 & 1e-6 & 0 & \tVal & 0.07644163 & 0.15745606 \\
        \fVal & 2000 & 6e-4 & 1e-7 & 0 & \tVal & 0.37478206 & 0.43952626 \\
        \fVal & 5000 & 1.5e-4 & 1e-7 & 0 & \fVal & 0.07908479 & 0.12611549 \\
        \tVal & 1000 & 1.5e-4 & 1e-7 & 0 & \fVal & 0.05360342 & 0.10210203 \\
        \tVal & 2000 & 5e-5 & 1e-7 & 0 & \fVal & 0.08639395 & 0.11047091 \\
        \tVal & 2000 & 1e-4 & 1e-7 & 0 & \fVal & 0.05374371 & 0.10861422 \\
        \tVal & 2000 & 1.5e-4 & 1e-8 & 0 & \fVal & 0.05400169 & 0.10814192 \\
        \tVal & 2000 & 1.5e-4 & 1e-7 & 0 & \fVal & 0.05369382 & 0.10388614 \\
        \tVal & 2000 & 1.5e-4 & 1e-7 & 0 & \fVal & 0.05369382 & 0.10388614 \\
        \tVal & 2000 & 1.5e-4 & 1e-7 & 0 & \fVal & 0.05369382 & 0.10388614 \\
        \tVal & 2000 & 1.5e-4 & 1e-7 & 0 & \fVal & 0.05369382 & 0.10388614 \\
        \tVal & 2000 & 1.5e-4 & 1e-7 & 0.1 & \fVal & 0.06541769 & 0.10765720 \\
        \tVal & 2000 & 1.5e-4 & 1e-6 & 0 & \fVal & 0.05431726 & 0.10708854 \\
        \tVal & 2000 & 6e-4 & 1e-7 & 0 & \tVal & NaN & NaN \\
        \tVal & 5000 & 1.5e-4 & 1e-7 & 0 & \fVal & 0.05561237 & 0.10878043 \\
    \end{tabular}
    \label{tab:raw_hyperparam_encoder_decoder}
\end{table}

\begin{table}[htbp]
    \centering
    \footnotesize
    \caption{Summarized results from the hyperparameter search for the encoder-decoder model. Individual results are shown in Table~\ref{tab:raw_hyperparam_encoder_decoder}. Chosen hyperparameters have been bolded.}
    \renewcommand{\arraystretch}{1.3}
  \begin{tabular}{@{} c c
        S[table-format=2.0, round-precision=0, table-text-alignment=center]
        S[table-format=2.0, round-precision=0, table-text-alignment=center]
        S[table-format=3.2, round-precision=2, table-text-alignment=center]
        S[table-format=1.4, round-precision=4, table-text-alignment=center]
        S[table-format=1.4, round-precision=4, table-text-alignment=center]
      @{}}

      & & \textbf{Runs} & \textbf{Divergences} & \textbf{Divergence Rate}~[$\%$] &
      \textbf{Training Loss} &
       \textbf{Validation Loss} \\
      \midrule
      \multirow[t]{2}{*}{\raisebox{-0.35cm}{\rotatebox{90}{CL}}}
          & Yes & 12 & 1 & 8.33 & 0.05799 & 0.10622 \rule{0pt}{0.5cm} \\[.1cm]
          & No  & 12 & 7 & 58.33 & 0.05882 & 0.11920  \\[.1cm]
      \midrule
      \addlinespace[0.6em]
      \multirow{3}{*}{\rotatebox{90}{Warmup\hspace{3pt}}}
          & 1000 & 2  & 0 & 0.00  & 0.05453 & 0.11745 \\
          & \bfseries 2000 & \bfseries 20 & \bfseries 8 & \bfseries 40.00 & \bfseries 0.05735 & \bfseries 0.10788 \\
          & 5000 & 2  & 0 & 0.00  & 0.06735 & 0.11745 \\
    \midrule
    \multirow{4}{*}{\rotatebox{90}{LR}}
        & 6.0~e-4   & 2  & 2 & 100.00 & {N/A} & {N/A} \\
        & 1.5~e-4 & 18 & 6 & 33.33 & 0.05712 & 0.11002 \\
        & \bfseries 1.0~e-4   & \bfseries 2  & \bfseries 0 & \bfseries 0.00  & \bfseries 0.04975 & \bfseries 0.10867 \\
        & 5.0~e-5   & 2  & 0 & 0.00  & 0.07349 & 0.11338 \\
      \midrule
      \multirow{3}{*}{\rotatebox{90}{Min LR}}
          & 1~e-6  & 2  & 1 & 50.00 & 0.05432 & 0.10709 \\
          & \bfseries 1~e-7  & \bfseries 20 & \bfseries 6 & \bfseries 30.00 & \bfseries 0.05883 & \bfseries 0.11065 \\
          & 1~e-8  & 2  & 1 & 50.00 & 0.05400 & 0.10814 \\
      \midrule
      \addlinespace[0.6em]
      \multirow[t]{2}{*}{\raisebox{-0.65cm}{\rotatebox{90}{Dropout}}}
          & 0   & 22 & 8 & 36.36 & 0.05810 & 0.11033  \rule{0pt}{0.4cm} \\[.08cm]
          & \bfseries 0.1 & \bfseries 2  & \bfseries 0 & \bfseries 0.00 \bfseries & \bfseries 0.05932 & \bfseries 0.10986 \\[.08cm]
    \end{tabular}
    \label{tab:summarized_hyperparam_encoder_decoder}
\end{table}

\subsection{Main Experiments}
\label{app:main-experiments}
For the main experiments, we used the hyperparameters from the hyperparameter search in Section~\ref{app:encoder-decoder-hyperparams} for the encoder and encoder-decoder architectures.
For this experiment, each epoch consisted of 100000 sampled states and plans and a batch size of 64 was used.
The maximum number of epochs was set to 70, with a timeout of 12 hours.
Additionally, the maximum number of objects was set to 123, a value slightly higher than the maximum number of objects in any test instance across all domains (121).
For each considered combination of domain, architecture and objective, we trained three models.
The same three seeds were used for each combination, meaning that the models with and without the contrastive loss started from the same initial weights.

During training, we scored the model on the validation test, saving the best performing model checkpoint. For this scoring, the following criteria were used (in decreasing order of importance):
\begin{itemize}
    \item Percentage of solved problems (coverage) with greedy plan generation
    \item Percentage of problems which reached the maximum number of tokens
    \item Percentage of problems where an invalid action was generated
    \item Percentage of problems where a malformed action was generated
    \item Accuracy between the predicted and sampled plans
    \item Loss
\end{itemize}
all computed on the validation set.
Though, note that the encoder architecture is unable to generate invalid or malformed actions.
The statistics such as coverage (Table~\ref{tab:coverage_split_base_woCL} and \ref{tab:coverage_split_inv_wCL}) and quality scores (Tables~\ref{tab:QS_split_base_wCL} - \ref{tab:QSS_split_inv_wCL}) are computed using these best checkpoints.

Tables \ref{tab:QS_split_base_wCL} - \ref{tab:QS_split_inv_wCL} and tables \ref{tab:QSS_split_base_wCL} - \ref{tab:QSS_split_inv_wCL} show the quality scores ($\text{QS}$) and the quality scores for solved problems ($\text{QS}_\text{S}$), respectively. These scores are computed as follows:
\begin{align}
 \text{QS} &= \frac {1} {|P|} \sum_{p \in P} \frac {|\pi^*(p)|} {|\pi(p)|} \\
\text{QS}_\text{S} &= \frac {1} {|\{p \in P: |\pi(p)| < \infty\}|} \sum_{p \in P} \frac {|\pi^*(p)|} {|\pi(p)|}
\end{align}
Where:
\begin{itemize}
    \item $P$ is the set of problems considered, such as the training, validation or either test set
    \item $\pi(p)$ is the predicted plan for problem $p$. An unsolved problem is assigned $|\pi(p)| = \infty$.
    \item $\pi^*(p)$ is the shortest known plan for problem $p$. We obtained these values with the state-of-the-art Scorpion planner \cite{seipp-et-al-jair2020}. For the few problems that could not be solved optimally by Scorpion  within 5\,hours and 64\,GiB, we used the suboptimal LAMA planner \cite{richter-westphal-jair2010} and a 2 hour limit to compute an upper bound on the optimal plan length instead.
\end{itemize}

For a total of 11 problems which Scorpion could not solve optimally, our encoder-decoder model with regrounding was able to find solutions shorter than those found by LAMA. Our most successful checkpoint outperformed LAMA across 5 problems. Across all models and problems where we outperformed LAMA, we on average reduced the plan length by $5.06\%$ ($4.44\%$ standard deviation) or $4.27$ ($4.76$ standard deviation) steps. In the best case, we reduced a plan length by $16.3\%$ (18~actions).

Additionally, we report the number of divergences for each architecture and domain in Table~\ref{tab:main-divergences}. No configuration using our contrastive loss diverged nor did any of the decoder-only models. For the encoder-only and encoder-decoder architectures without our contrastive loss, we observed a total of 9 ($75\%$) and 3 ($25\%$) divergences, respectively. Notably, no divergences ever occur in the Gripper domain.

In total, we trained 72 models each with a timeout of 12 hours, using 1440 hours of GPU time for this experiment. %

  \setlength{\tabcolsep}{2pt}
  \begin{table*}[tb]
    \renewcommand{\arraystretch}{1.1}
    \centering
    \resizebox{\textwidth}{!}{
      \newcolumntype{Y}{>{\centering\arraybackslash}X}
 \begin{tabularx}{\textwidth}{clYYYYYYY}
 & & \multicolumn{3}{c}{PlanGPT - Decoder (baseline) ~w/~CL} & \multicolumn{1}{c}{$\text{SymT}^{\text{E}}$ (ours) w/o~CL} & \multicolumn{3}{c}{$\text{SymT}^{\text{ED}}$ (ours) w/o~CL} \\
 \cmidrule(lr){3-5} \cmidrule(lr){6-6} \cmidrule(lr){7-9}
 & & greedy & applicable & regrounding & \hspace*{3.25em} greedy & greedy & applicable & regrounding \\
 \midrule
 \multirow{3}{*}{\rotatebox[origin=c]{90}{Blocks}} & validation & ~\,.00$\pm$.00 & ~\,.00$\pm$.00 & ~\,.00$\pm$.00 & \hspace*{2.9em} ~\,.00$\pm$.00 & \textbf{1.00$\pm$.00} & \textbf{1.00$\pm$.00} & ~\,.00$\pm$.00 \\
 & interpolation & ~\,.67$\pm$.00 & ~\,.67$\pm$.00 & ~\,.00$\pm$.00 & \hspace*{2.9em} ~\,.00$\pm$.00 & \textbf{1.00$\pm$.00} & \textbf{1.00$\pm$.00} & \textbf{1.00$\pm$.00} \\
 & extrapolation & ~\,.00$\pm$.00 & ~\,.00$\pm$.00 & ~\,.00$\pm$.00 & \hspace*{2.9em} ~\,.00$\pm$.00 & ~\,.03$\pm$.02 & \textbf{~\,.15$\pm$.07} & ~\,.00$\pm$.00 \\
 \midrule
 \multirow{3}{*}{\rotatebox[origin=c]{90}{Gripper}} & validation & ~\,.00$\pm$.00 & ~\,.00$\pm$.00 & ~\,.00$\pm$.00 & \hspace*{2.75em} \textbf{1.00$\pm$.00} & ~\,.00$\pm$.00 & \textbf{1.00$\pm$.00} & \textbf{1.00$\pm$.00} \\
 & interpolation & ~\,.00$\pm$.00 & ~\,.11$\pm$.16 & ~\,.00$\pm$.00 & \hspace*{2.75em} \textbf{1.00$\pm$.00} & ~\,.78$\pm$.16 & \textbf{1.00$\pm$.00} & \textbf{1.00$\pm$.00} \\
 & extrapolation & ~\,.00$\pm$.00 & ~\,.00$\pm$.00 & ~\,.00$\pm$.00 & \hspace*{2.9em} ~\,.04$\pm$.03 & ~\,.00$\pm$.00 & ~\,.10$\pm$.03 & \textbf{~\,.85$\pm$.13} \\
 \midrule
 \multirow{3}{*}{\rotatebox[origin=c]{90}{Visitall}} & validation & ~\,.00$\pm$.00 & ~\,.42$\pm$.12 & ~\,.00$\pm$.00 & \hspace*{2.9em} ~\,.42$\pm$.03 & ~\,.15$\pm$.05 & ~\,.65$\pm$.17 & \textbf{~\,.96$\pm$.06} \\
 & interpolation & ~\,.04$\pm$.03 & ~\,.86$\pm$.08 & ~\,.50$\pm$.07 & \hspace*{2.9em} ~\,.80$\pm$.09 & ~\,.85$\pm$.03 & ~\,.98$\pm$.01 & \textbf{1.00$\pm$.00} \\
 & extrapolation & ~\,.00$\pm$.00 & ~\,.04$\pm$.02 & ~\,.00$\pm$.00 & \hspace*{2.9em} ~\,.04$\pm$.00 & ~\,.00$\pm$.00 & ~\,.05$\pm$.00 & \textbf{~\,.47$\pm$.14} \\
 \midrule
 \multirow{3}{*}{\rotatebox[origin=c]{90}{Logistics}} & validation & \textbf{~\,.00$\pm$.00} & \textbf{~\,.00$\pm$.00} & \textbf{~\,.00$\pm$.00} & \hspace*{2.9em} \textbf{~\,.00$\pm$.00} & \textbf{~\,.00$\pm$.00} & \textbf{~\,.00$\pm$.00} & \textbf{~\,.00$\pm$.00} \\
 & interpolation & ~\,.15$\pm$.14 & \textbf{~\,.30$\pm$.23} & ~\,.22$\pm$.18 & \hspace*{2.9em} ~\,.04$\pm$.05 & ~\,.11$\pm$.16 & ~\,.19$\pm$.10 & ~\,.04$\pm$.05 \\
 & extrapolation & \textbf{~\,.00$\pm$.00} & \textbf{~\,.00$\pm$.00} & \textbf{~\,.00$\pm$.00} & \hspace*{2.9em} \textbf{~\,.00$\pm$.00} & \textbf{~\,.00$\pm$.00} & \textbf{~\,.00$\pm$.00} & \textbf{~\,.00$\pm$.00} \\
 \bottomrule
 \end{tabularx}
    }
    \caption{Normalized \textbf{coverage scores} $ [(\mu \pm \sigma)] $ for all \textbf{ablation models} (contrastive loss inverted) and plan generation strategies. A coverage score of 1.00 means that the configuration solves all tasks.}
    \label{tab:coverage_split_inv_wCL}
  \end{table*}

  \setlength{\tabcolsep}{2pt}
  \begin{table*}[tb]
    \renewcommand{\arraystretch}{1.1}
    \centering
    \resizebox{\textwidth}{!}{
      \newcolumntype{Y}{>{\centering\arraybackslash}X}
 \begin{tabularx}{\textwidth}{clYYYYYYY}
 & & \multicolumn{3}{c}{PlanGPT - Decoder (baseline) ~w/o~CL} & \multicolumn{1}{c}{$\text{SymT}^{\text{E}}$ (ours) w/~CL} & \multicolumn{3}{c}{$\text{SymT}^{\text{ED}}$ (ours) w/~CL} \\
 \cmidrule(lr){3-5} \cmidrule(lr){6-6} \cmidrule(lr){7-9}
 & & greedy & applicable & regrounding & \hspace*{3.25em} greedy & greedy & applicable & regrounding \\
 \midrule
 \multirow{3}{*}{\rotatebox[origin=c]{90}{Blocks}} & validation & ~\,.00$\pm$.00 & ~\,.00$\pm$.00 & ~\,.00$\pm$.00 & \hspace*{2.75em} \textbf{1.00$\pm$.00} & \textbf{1.00$\pm$.00} & \textbf{1.00$\pm$.00} & ~\,.00$\pm$.00 \\
 & interpolation & ~\,.56$\pm$.16 & ~\,.56$\pm$.16 & ~\,.00$\pm$.00 & \hspace*{2.75em} \textbf{1.00$\pm$.00} & \textbf{1.00$\pm$.00} & \textbf{1.00$\pm$.00} & \textbf{1.00$\pm$.00} \\
 & extrapolation & ~\,.00$\pm$.00 & ~\,.00$\pm$.00 & ~\,.00$\pm$.00 & \hspace*{2.9em} ~\,.05$\pm$.07 & ~\,.07$\pm$.02 & \textbf{~\,.13$\pm$.05} & ~\,.00$\pm$.00 \\
 \midrule
 \multirow{3}{*}{\rotatebox[origin=c]{90}{Gripper}} & validation & ~\,.00$\pm$.00 & ~\,.00$\pm$.00 & ~\,.00$\pm$.00 & \hspace*{2.9em} \textbf{~\,.99$\pm$.02} & ~\,.17$\pm$.24 & ~\,.98$\pm$.03 & \textbf{~\,.99$\pm$.02} \\
 & interpolation & ~\,.00$\pm$.00 & ~\,.36$\pm$.06 & ~\,.00$\pm$.00 & \hspace*{2.9em} ~\,.89$\pm$.16 & ~\,.67$\pm$.00 & \textbf{1.00$\pm$.00} & \textbf{1.00$\pm$.00} \\
 & extrapolation & ~\,.00$\pm$.00 & ~\,.00$\pm$.00 & ~\,.00$\pm$.00 & \hspace*{2.9em} ~\,.02$\pm$.03 & ~\,.00$\pm$.00 & ~\,.14$\pm$.06 & \textbf{~\,.77$\pm$.14} \\
 \midrule
 \multirow{3}{*}{\rotatebox[origin=c]{90}{Visitall}} & validation & ~\,.00$\pm$.00 & ~\,.07$\pm$.05 & ~\,.00$\pm$.00 & \hspace*{2.75em} \textbf{1.00$\pm$.00} & ~\,.32$\pm$.10 & ~\,.76$\pm$.04 & ~\,.96$\pm$.02 \\
 & interpolation & ~\,.05$\pm$.04 & ~\,.57$\pm$.15 & ~\,.41$\pm$.21 & \hspace*{2.75em} \textbf{1.00$\pm$.00} & ~\,.87$\pm$.02 & ~\,.96$\pm$.01 & 1.00$\pm$.00 \\
 & extrapolation & ~\,.00$\pm$.00 & ~\,.01$\pm$.01 & ~\,.00$\pm$.00 & \hspace*{2.9em} ~\,.40$\pm$.09 & ~\,.00$\pm$.00 & ~\,.10$\pm$.03 & \textbf{~\,.60$\pm$.12} \\
 \midrule
 \multirow{3}{*}{\rotatebox[origin=c]{90}{Logistics}} & validation & ~\,.00$\pm$.00 & \textbf{~\,.08$\pm$.12} & ~\,.00$\pm$.00 & \hspace*{2.9em} ~\,.00$\pm$.00 & ~\,.00$\pm$.00 & ~\,.00$\pm$.00 & ~\,.00$\pm$.00 \\
 & interpolation & ~\,.07$\pm$.05 & \textbf{~\,.38$\pm$.04} & ~\,.19$\pm$.14 & \hspace*{2.9em} ~\,.11$\pm$.00 & ~\,.22$\pm$.31 & ~\,.24$\pm$.30 & ~\,.22$\pm$.31 \\
 & extrapolation & \textbf{~\,.00$\pm$.00} & \textbf{~\,.00$\pm$.00} & \textbf{~\,.00$\pm$.00} & \hspace*{2.9em} \textbf{~\,.00$\pm$.00} & \textbf{~\,.00$\pm$.00} & \textbf{~\,.00$\pm$.00} & \textbf{~\,.00$\pm$.00} \\
 \bottomrule
 \end{tabularx}
    }
    \caption{Normalized \textbf{quality scores} ($\text{QS}$) $ [(\mu \pm \sigma)] $ for all \textbf{base models} and plan generation strategies. A quality score score of 1.00 means that the configuration solves all problems optimally. The best results are highlighted in bold.}
    \label{tab:QS_split_base_wCL}
  \end{table*}

  \setlength{\tabcolsep}{2pt}
  \begin{table*}[tb]
    \renewcommand{\arraystretch}{1.1}
    \centering
    \resizebox{\textwidth}{!}{
      \newcolumntype{Y}{>{\centering\arraybackslash}X}
 \begin{tabularx}{\textwidth}{clYYYYYYY}
 & & \multicolumn{3}{c}{PlanGPT - Decoder (baseline) ~w/~CL} & \multicolumn{1}{c}{$\text{SymT}^{\text{E}}$ (ours) w/o~CL} & \multicolumn{3}{c}{$\text{SymT}^{\text{ED}}$ (ours) w/o~CL} \\
 \cmidrule(lr){3-5} \cmidrule(lr){6-6} \cmidrule(lr){7-9}
 & & greedy & applicable & regrounding & \hspace*{3.25em} greedy & greedy & applicable & regrounding \\
 \midrule
 \multirow{3}{*}{\rotatebox[origin=c]{90}{Blocks}} & validation & ~\,.00$\pm$.00 & ~\,.00$\pm$.00 & ~\,.00$\pm$.00 & \hspace*{2.9em} ~\,.00$\pm$.00 & \textbf{1.00$\pm$.00} & \textbf{1.00$\pm$.00} & ~\,.00$\pm$.00 \\
 & interpolation & ~\,.67$\pm$.00 & ~\,.67$\pm$.00 & ~\,.00$\pm$.00 & \hspace*{2.9em} ~\,.00$\pm$.00 & \textbf{1.00$\pm$.00} & \textbf{1.00$\pm$.00} & \textbf{1.00$\pm$.00} \\
 & extrapolation & ~\,.00$\pm$.00 & ~\,.00$\pm$.00 & ~\,.00$\pm$.00 & \hspace*{2.9em} ~\,.00$\pm$.00 & ~\,.03$\pm$.02 & \textbf{~\,.15$\pm$.07} & ~\,.00$\pm$.00 \\
 \midrule
 \multirow{3}{*}{\rotatebox[origin=c]{90}{Gripper}} & validation & ~\,.00$\pm$.00 & ~\,.00$\pm$.00 & ~\,.00$\pm$.00 & \hspace*{2.75em} \textbf{1.00$\pm$.00} & ~\,.00$\pm$.00 & ~\,.99$\pm$.02 & \textbf{1.00$\pm$.00} \\
 & interpolation & ~\,.00$\pm$.00 & ~\,.10$\pm$.14 & ~\,.00$\pm$.00 & \hspace*{2.75em} \textbf{1.00$\pm$.00} & ~\,.78$\pm$.16 & \textbf{1.00$\pm$.00} & \textbf{1.00$\pm$.00} \\
 & extrapolation & ~\,.00$\pm$.00 & ~\,.00$\pm$.00 & ~\,.00$\pm$.00 & \hspace*{2.9em} ~\,.04$\pm$.03 & ~\,.00$\pm$.00 & ~\,.09$\pm$.02 & \textbf{~\,.84$\pm$.12} \\
 \midrule
 \multirow{3}{*}{\rotatebox[origin=c]{90}{Visitall}} & validation & ~\,.00$\pm$.00 & ~\,.24$\pm$.07 & ~\,.00$\pm$.00 & \hspace*{2.9em} ~\,.38$\pm$.03 & ~\,.15$\pm$.05 & ~\,.50$\pm$.12 & \textbf{~\,.91$\pm$.04} \\
 & interpolation & ~\,.04$\pm$.03 & ~\,.76$\pm$.07 & ~\,.50$\pm$.07 & \hspace*{2.9em} ~\,.77$\pm$.09 & ~\,.82$\pm$.03 & ~\,.95$\pm$.01 & \textbf{~\,.99$\pm$.00} \\
 & extrapolation & ~\,.00$\pm$.00 & ~\,.02$\pm$.01 & ~\,.00$\pm$.00 & \hspace*{2.9em} ~\,.03$\pm$.00 & ~\,.00$\pm$.00 & ~\,.03$\pm$.00 & \textbf{~\,.43$\pm$.13} \\
 \midrule
 \multirow{3}{*}{\rotatebox[origin=c]{90}{Logistics}} & validation & \textbf{~\,.00$\pm$.00} & \textbf{~\,.00$\pm$.00} & \textbf{~\,.00$\pm$.00} & \hspace*{2.9em} \textbf{~\,.00$\pm$.00} & \textbf{~\,.00$\pm$.00} & \textbf{~\,.00$\pm$.00} & \textbf{~\,.00$\pm$.00} \\
 & interpolation & ~\,.15$\pm$.14 & \textbf{~\,.27$\pm$.21} & ~\,.22$\pm$.18 & \hspace*{2.9em} ~\,.04$\pm$.05 & ~\,.11$\pm$.16 & ~\,.12$\pm$.15 & ~\,.04$\pm$.05 \\
 & extrapolation & \textbf{~\,.00$\pm$.00} & \textbf{~\,.00$\pm$.00} & \textbf{~\,.00$\pm$.00} & \hspace*{2.9em} \textbf{~\,.00$\pm$.00} & \textbf{~\,.00$\pm$.00} & \textbf{~\,.00$\pm$.00} & \textbf{~\,.00$\pm$.00} \\
 \bottomrule
 \end{tabularx}
    }
    \caption{Normalized \textbf{quality scores} ($\text{QS}$) $ [(\mu \pm \sigma)] $ for all \textbf{ablation models} (contrastive loss inverted) and plan generation strategies. A quality score score of 1.00 means that the configuration solves all problems optimally. The best results are highlighted in bold.}
    \label{tab:QS_split_inv_wCL}
  \end{table*}

  \setlength{\tabcolsep}{2pt}
  \begin{table*}[tb]
    \renewcommand{\arraystretch}{1.1}
    \centering
    \resizebox{\textwidth}{!}{
      \newcolumntype{Y}{>{\centering\arraybackslash}X}
 \begin{tabularx}{\textwidth}{clYYYYYYY}
 & & \multicolumn{3}{c}{PlanGPT - Decoder (baseline) ~w/o~CL} & \multicolumn{1}{c}{$\text{SymT}^{\text{E}}$ (ours) w/~CL} & \multicolumn{3}{c}{$\text{SymT}^{\text{ED}}$ (ours) w/~CL} \\
 \cmidrule(lr){3-5} \cmidrule(lr){6-6} \cmidrule(lr){7-9}
 & & greedy & applicable & regrounding & \hspace*{3.25em} greedy & greedy & applicable & regrounding \\
 \midrule
 \multirow{3}{*}{\rotatebox[origin=c]{90}{Blocks}} & validation & \NA & \NA & \NA & \hspace*{2.75em} \textbf{1.00$\pm$.00} & \textbf{1.00$\pm$.00} & \textbf{1.00$\pm$.00} & \NA \\
 & interpolation & \textbf{1.00$\pm$.00} & \textbf{1.00$\pm$.00} & \NA & \hspace*{2.75em} \textbf{1.00$\pm$.00} & \textbf{1.00$\pm$.00} & \textbf{1.00$\pm$.00} & \textbf{1.00$\pm$.00} \\
 & extrapolation & \NA & \NA & \NA & \hspace*{2.75em} \textbf{1.00$\pm$.00} & \textbf{1.00$\pm$.00} & \textbf{1.00$\pm$.00} & \NA \\
 \midrule
 \multirow{3}{*}{\rotatebox[origin=c]{90}{Gripper}} & validation & \NA & \NA & \NA & \hspace*{2.9em} ~\,.99$\pm$.02 & \textbf{1.00$\pm$.00} & ~\,.98$\pm$.03 & ~\,.99$\pm$.02 \\
 & interpolation & \NA & ~\,.85$\pm$.14 & \NA & \hspace*{2.75em} \textbf{1.00$\pm$.00} & \textbf{1.00$\pm$.00} & \textbf{1.00$\pm$.00} & \textbf{1.00$\pm$.00} \\
 & extrapolation & \NA & \NA & \NA & \hspace*{2.9em} ~\,.95$\pm$.00 & \NA & ~\,.95$\pm$.02 & \textbf{~\,.97$\pm$.02} \\
 \midrule
 \multirow{3}{*}{\rotatebox[origin=c]{90}{Visitall}} & validation & \NA & ~\,.51$\pm$.08 & \NA & \hspace*{2.75em} \textbf{1.00$\pm$.00} & ~\,.93$\pm$.06 & ~\,.82$\pm$.06 & ~\,.97$\pm$.01 \\
 & interpolation & \textbf{1.00$\pm$.00} & ~\,.86$\pm$.06 & ~\,.99$\pm$.01 & \hspace*{2.75em} \textbf{1.00$\pm$.00} & ~\,.99$\pm$.01 & ~\,.97$\pm$.00 & 1.00$\pm$.00 \\
 & extrapolation & \NA & ~\,.42$\pm$.00 & \NA & \hspace*{2.9em} \textbf{~\,.98$\pm$.00} & ~\,.96$\pm$.04 & ~\,.63$\pm$.01 & ~\,.93$\pm$.02 \\
 \midrule
 \multirow{3}{*}{\rotatebox[origin=c]{90}{Logistics}} & validation & \NA & \textbf{1.00$\pm$.00} & \NA & \hspace*{2.75em} \NA & \NA & \NA & \NA \\
 & interpolation & \textbf{1.00$\pm$.00} & ~\,.89$\pm$.12 & \textbf{1.00$\pm$.00} & \hspace*{2.75em} \textbf{1.00$\pm$.00} & \textbf{1.00$\pm$.00} & ~\,.69$\pm$.31 & \textbf{1.00$\pm$.00} \\
 & extrapolation & \NA & \NA & \NA & \hspace*{2.75em} \NA & \NA & \NA & \NA \\
 \bottomrule
 \end{tabularx}
    }
    \caption{Normalized \textbf{quality scores for solved problems} ($\text{QS}_{\text{S}}$) $ [(\mu \pm \sigma)] $ for all \textbf{base models} and plan generation strategies. A quality score for solved problems of 1.00 means that for every problem the configuration solves, it does so optimally. If the configuration failed to solve any problem, we report it as \NA. The best results are highlighted in bold.}
    \label{tab:QSS_split_base_wCL}
  \end{table*}

  \setlength{\tabcolsep}{2pt}
  \begin{table*}[tb]
    \renewcommand{\arraystretch}{1.1}
    \centering
    \resizebox{\textwidth}{!}{
      \newcolumntype{Y}{>{\centering\arraybackslash}X}
 \begin{tabularx}{\textwidth}{clYYYYYYY}
 & & \multicolumn{3}{c}{PlanGPT - Decoder (baseline) ~w/~CL} & \multicolumn{1}{c}{$\text{SymT}^{\text{E}}$ (ours) w/o~CL} & \multicolumn{3}{c}{$\text{SymT}^{\text{ED}}$ (ours) w/o~CL} \\
 \cmidrule(lr){3-5} \cmidrule(lr){6-6} \cmidrule(lr){7-9}
 & & greedy & applicable & regrounding & \hspace*{3.25em} greedy & greedy & applicable & regrounding \\
 \midrule
 \multirow{3}{*}{\rotatebox[origin=c]{90}{Blocks}} & validation & \NA & \NA & \NA & \hspace*{2.75em} \NA & \textbf{1.00$\pm$.00} & \textbf{1.00$\pm$.00} & \NA \\
 & interpolation & \textbf{1.00$\pm$.00} & \textbf{1.00$\pm$.00} & \NA & \hspace*{2.75em} \NA & \textbf{1.00$\pm$.00} & \textbf{1.00$\pm$.00} & \textbf{1.00$\pm$.00} \\
 & extrapolation & \NA & \NA & \NA & \hspace*{2.75em} \NA & \textbf{1.00$\pm$.00} & \textbf{1.00$\pm$.00} & \NA \\
 \midrule
 \multirow{3}{*}{\rotatebox[origin=c]{90}{Gripper}} & validation & \NA & \NA & \NA & \hspace*{2.75em} \textbf{1.00$\pm$.00} & \NA & ~\,.99$\pm$.02 & \textbf{1.00$\pm$.00} \\
 & interpolation & \NA & ~\,.88$\pm$.00 & \NA & \hspace*{2.75em} \textbf{1.00$\pm$.00} & \textbf{1.00$\pm$.00} & \textbf{1.00$\pm$.00} & \textbf{1.00$\pm$.00} \\
 & extrapolation & \NA & \NA & \NA & \hspace*{2.9em} ~\,.95$\pm$.00 & \NA & ~\,.91$\pm$.09 & \textbf{~\,.99$\pm$.00} \\
 \midrule
 \multirow{3}{*}{\rotatebox[origin=c]{90}{Visitall}} & validation & \NA & ~\,.57$\pm$.03 & \NA & \hspace*{2.9em} ~\,.90$\pm$.02 & \textbf{~\,.98$\pm$.01} & ~\,.78$\pm$.03 & ~\,.95$\pm$.02 \\
 & interpolation & \textbf{1.00$\pm$.00} & ~\,.88$\pm$.01 & 1.00$\pm$.00 & \hspace*{2.9em} ~\,.96$\pm$.00 & ~\,.97$\pm$.02 & ~\,.97$\pm$.02 & ~\,.99$\pm$.00 \\
 & extrapolation & \NA & ~\,.48$\pm$.02 & \NA & \hspace*{2.9em} ~\,.88$\pm$.01 & \textbf{1.00$\pm$.00} & ~\,.62$\pm$.05 & ~\,.92$\pm$.02 \\
 \midrule
 \multirow{3}{*}{\rotatebox[origin=c]{90}{Logistics}} & validation & \NA & \NA & \NA & \hspace*{2.75em} \NA & \NA & \NA & \NA \\
 & interpolation & \textbf{1.00$\pm$.00} & ~\,.92$\pm$.00 & \textbf{1.00$\pm$.00} & \hspace*{2.75em} \textbf{1.00$\pm$.00} & \textbf{1.00$\pm$.00} & ~\,.43$\pm$.40 & \textbf{1.00$\pm$.00} \\
 & extrapolation & \NA & \NA & \NA & \hspace*{2.75em} \NA & \NA & \NA & \NA \\
 \bottomrule
 \end{tabularx}
    }
    \caption{Normalized \textbf{quality scores for solved problems} ($\text{QS}_{\text{S}}$) $ [(\mu \pm \sigma)] $ for all \textbf{ablation models} (contrastive loss inverted) and plan generation strategies. A quality score for solved problems of 1.00 means that for every problem the configuration solves, it does so optimally. If the configuration failed to solve any problem, we report it as \NA. The best results are highlighted in bold.}
    \label{tab:QSS_split_inv_wCL}
  \end{table*}

\begin{table}[htbp]
    \centering
    \caption{Number of divergences for all architectures and domains for the main experiments. We trained three models per configuration.}
    \begin{tabular}{@{}l cc  cc cc@{}}
        & \multicolumn{2}{c}{PlanGPT} & \multicolumn{2}{c}{$\text{SymT}^{\text{E}}$} & \multicolumn{2}{c}{$\text{SymT}^{\text{ED}}$} \\
        \cmidrule(lr){2-3} \cmidrule(lr){4-5} \cmidrule(lr){6-7}
        & No CL & CL & No CL & CL & No CL & CL \\
        \midrule
        Blocksworld
            & 0 & 0 & 3 & 0 & 1 & 0 \\
        Gripper
            & 0 & 0 & 0 & 0 & 0 & 0 \\
        Logistics
            & 0 & 0 & 3 & 0 & 1 & 0 \\
        Visitall
            & 0 & 0 & 3 & 0 & 1 & 0 \\
    \end{tabular}
    \label{tab:main-divergences}
\end{table}

\clearpage
\subsection{Effect of Number of Objects on Contrastive Loss}
\label{app:contrastive-learning-objective-num-objects}
The primary motivation for employing a contrastive learning objective is to mitigate the combinatorial explosion inherent in object assignments within planning problems.
As such, we hypothesize that the objective's effectiveness will be most pronounced when the number of potential object names is large.

\begin{figure}[t]
    \centering
    \includegraphics[width=0.8\textwidth]{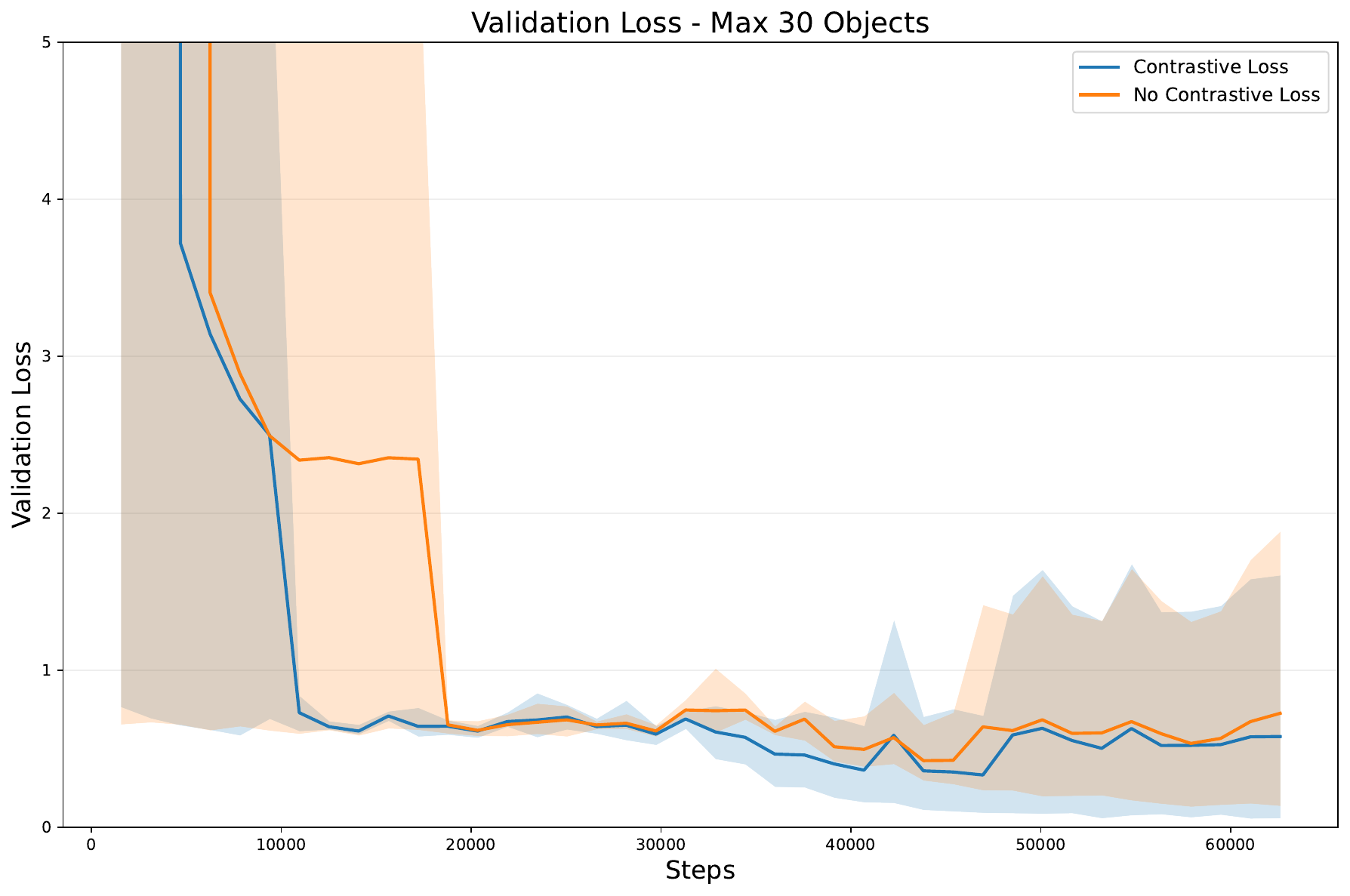}
    \caption{Validation loss curves for the Visitall domain when supporting 30 objects, comparing encoder-only models with (blue) and without (orange) our contrastive learning objective. We trained three models per configuration. The line shows the mean and the shaded region shows the maximum and minimum loss across these three models.}
    \label{fig:30-obj-visitall-append}
\end{figure}

\begin{figure}[t]
    \centering
    \includegraphics[width=0.8\textwidth]{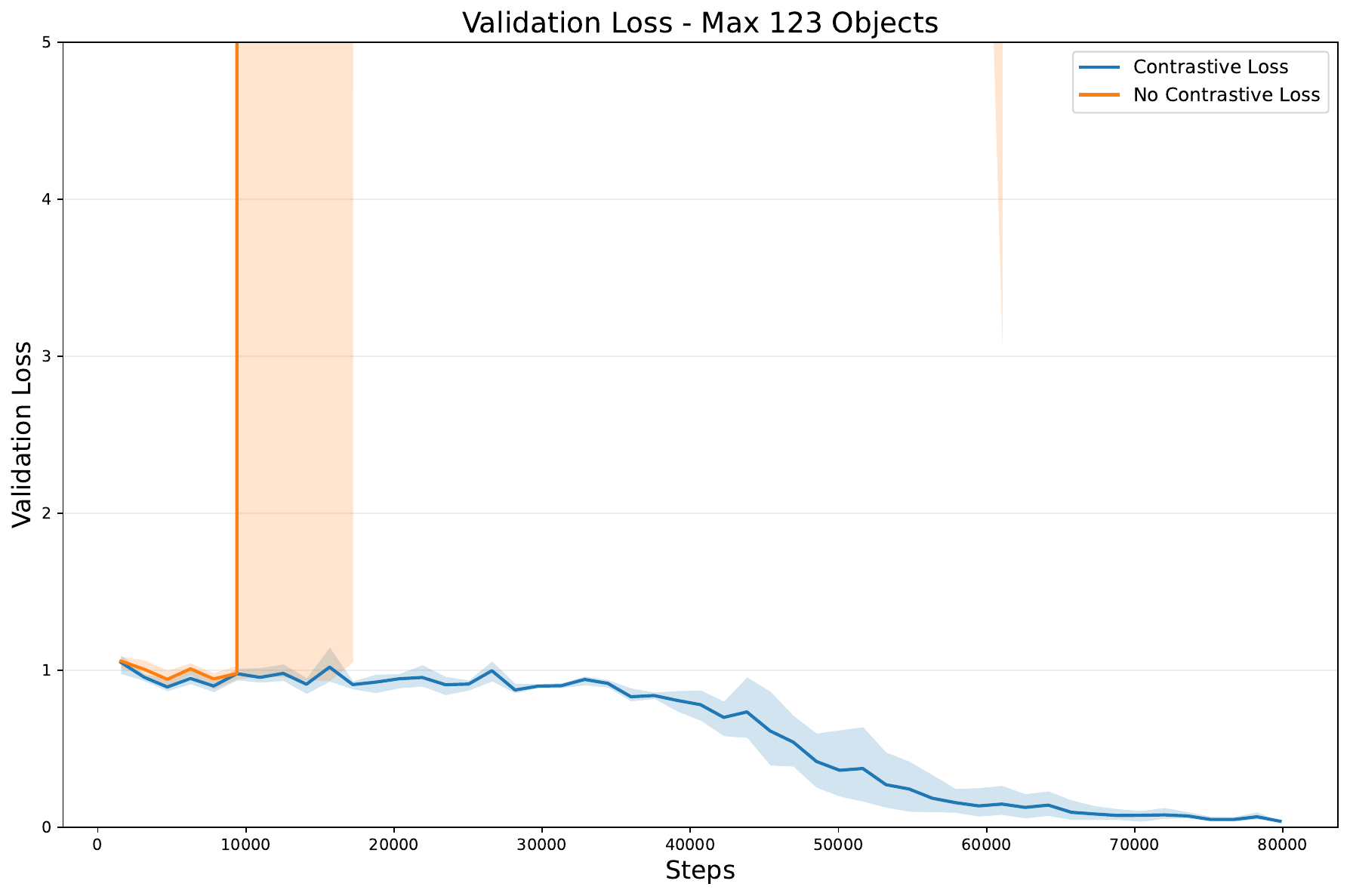}
    \caption{Validation loss curves for the Visitall domain when supporting 123 objects, comparing encoder-only models with (blue) and without (orange) our contrastive learning objective. We trained three models per configuration. The line shows the mean and the shaded region shows the maximum and minimum loss across these three models.}
    \label{fig:123-obj-visitall-append}
\end{figure}

To investigate this, we examine training loss curves for the encoder-only architecture in the Visitall domain with varying numbers of objects. Figure~\ref{fig:30-obj-visitall-append} illustrates results for the 30-object case (using Rename-One mode for object renaming), where both models with and without the contrastive objective converge to low loss values.
However, the model incorporating the contrastive objective exhibits a lower final loss and faster convergence compared to the baseline without it on average.
The benefits become significantly more apparent in the 123-object case (Figure~\ref{fig:123-obj-visitall-append}, also "Rename-One" mode). Here, the model with the contrastive learning objective successfully converges to a low loss, whereas models without the objective reach a plateau and fail to converge within the training duration. We observed similar behavior across other planning domains evaluated.

An additional factor influencing performance is the choice of renaming mode.
Specifically, preliminary experiments showed that applying the "Rename-Both" mode generally prevented encoder-only models from converging to a low loss.
In this challenging setting, the contrastive learning objective again demonstrated its efficacy, enabling convergence to a low loss in one of the experimental runs, a result not achieved by models without the objective in this mode.

For this experiment, we re-used the 123-object models trained in the main experiments, and as such only had to train the 30-object models. This resulted in a total of 6 models trained, using 72 hours of total GPU time (12 hours per model).

\subsection{Preliminary Experiments}
A variety of preliminary experiments were also conducted to debug our implementation and to explore the impact of different design choices. However, note that we did not evaluate any models trained in these experiments on the test set. These experiments were less accurately tracked and we are unaware of the exact amount of GPU time used for them.

\end{document}